\documentclass{article} 
\usepackage{iclr2022_conference,times}

\usepackage[utf8]{inputenc} 
\usepackage[T1]{fontenc}    
\usepackage{hyperref}       
\usepackage{url}            
\usepackage{booktabs}       
\usepackage{amsfonts}       
\usepackage{nicefrac}       
\usepackage{xcolor}         
\usepackage{caption}
\usepackage{subcaption}
\usepackage[paperVersion]{dpupaper}

\newcommand{\projName}{\textsc{Grammformer}\xspace}
\newcommand{\hole}{\ensuremath{\blacksquare}\xspace}
\newcommand{\NT}[1]{\ensuremath{\left\langle\textsf{#1}\right\rangle}}

\newcommand{\inSeq}[0]{\ensuremath{\mathbf{x}}\xspace}
\newcommand{\expSeq}[0]{\ensuremath{\mathbf{u}}\xspace}
\newcommand{\outSeq}[0]{\ensuremath{\mathbf{y}}\xspace}
\newcommand{\pred}[1]{\ensuremath{\hat{#1}}\xspace}
\newcommand{\gt}[1]{\ensuremath{#1^\ast}\xspace}
\newcommand{\regexAcc}{\textsc{RegexAcc}\xspace}
\newcommand{\rouge}{\textsc{Rouge}\xspace}
\newcommand{\selectorModel}[0]{\ensuremath{P_s}\xspace}
\newcommand{\expansionModel}[0]{\ensuremath{P_e}\xspace}
\newcommand{\stopExpand}{\ensuremath{{\color{red}\obslash}}\xspace}
\newcommand{\lrbaseline}{\ensuremath{L\rightarrow R}\xspace}
\newcommand{\lrstopbaseline}{\ensuremath{L\rightarrow R + \stopExpand}\xspace}
\newcommand{\lrholebaseline}{\ensuremath{L\rightarrow R + \hole}\xspace}
\newcommand{\lmbaseline}{\ensuremath{LM}\xspace}
\newcommand{\lmstopbaseline}{\ensuremath{LM + \stopExpand}\xspace}
\newcommand{\lmholebaseline}{\ensuremath{LM + \hole}\xspace}
\newcommand{\lmc}{LMC\xspace}
\newcommand{\lmcs}{LMCs\xspace}

\title{Learning to Complete Code with Sketches}

\author{%
  Daya Guo\\
  Microsoft Research\\
  Beijing, China\\
  \texttt{t-dayaguo@microsoft.com} \\
  \And
  Alexey Svyatkovskiy \\
  Microsoft \\
  Redmond, WA, USA\\
  \texttt{alsvyatk@microsoft.com}
  \And
  Jian Yin \\
  School of Data and Computer Science\\ Sun Yat-sen University, China\\
  \texttt{issjyin@mail.sysu.edu.cn}
  \And
  Nan Duan \\
  Microsoft Research \\
  Beijing, China\\
  \texttt{nanduan@microsoft.com}
  \AND
  Marc Brockschmidt, Miltiadis Allamanis \\
  Microsoft Research \\
  Cambridge, UK \\
  \texttt{\{mabrocks,miallama\}@microsoft.com} \\
}

\iclrfinalcopy
\begin{document}

\maketitle

\begin{abstract}
Code completion is usually cast as a language modelling problem, \ie, continuing an input in a left-to-right fashion.
However, in practice, some parts of the completion (\eg, string literals) may be very hard to predict, whereas subsequent parts directly follow from the context.
To handle this, we instead consider the scenario of generating code completions with ``holes'' inserted in places where a model is uncertain.
We develop \projName, a Transformer-based model that guides code generation by the programming language grammar, and compare it to a variety of more standard sequence models.

We train the models on code completion for C\# and Python given partial code context.
To evaluate models, we consider both \rouge as well as a new metric \regexAcc that measures success of generating completions matching long outputs with as few holes as possible.
In our experiments, \projName generates 10-50\% more accurate completions compared to traditional generative models and 37-50\% longer sketches compared to sketch-generating baselines trained with similar techniques.
\end{abstract}

\section{Introduction}
\label{sec:intro}
Recent high-capacity language models (LM) have shown that machine learning models are able to generate coherent, realistic text, but it is often hard to guide them towards a specific goal, especially when describing the intent is complex or more costly than manually generating the target output.

One such scenario are LMs of source code (\lmc). Since \citet{hindle2012naturalness} increasingly sophisticated \lmcs have been built, including transformer-based ones, such as those of \citet{svyatkovskiy2020intellicode,feng2020codebert,chen2021evaluating} and various similar unpublished models such as \href{https://www.tabnine.com/}{TabNine} and \href{https://sourceai.dev/}{SourceAI}.
These models generate full sequences of code tokens left-to-right with any prefix acting as the (partial) user intent.
While LMs generate realistic-looking outputs, they are known to occasionally ``hallucinate''~\citep{puduppully2019data,malmi2019encode,maynez2020faithfulness,liu2021token}, \ie generate plausible but incorrect content.
This is particularly problematic in generating source code, where small mistakes can lead to erroneous code that is very hard to debug or introduces vulnerabilities~\citep{pearce2021empirical}.

In this work, we investigate models that can decline to make predictions in places where there is high uncertainty (\eg, where the user should choose a name), but continue generating around these ``holes''.
For example, in \autoref{fig:qual example}(left) a developer has typed some code and is about to type the next line.
A likely completion is to consume more command line arguments, but their name is unclear from the context.
A traditional generative model (\eg \autoref{fig:qual example}; top right) may choose to provide a completion that exists in the training data, but is not clearly called for here.
On the other hand, a model able to explicitly mark where it is uncertain (\autoref{fig:qual example}; bottom right) makes it clear to a user where further input is required.

However, creating such models is not trivial.
A simple first attempt may be to use a standard \lmc, but output a ``hole token'' \hole whenever the model is uncertain about the next output token.
However, continuing after the ``\hole'' then becomes infeasible, as the \lmc was not trained on such data.
Hence, a suitable training dataset and objective need to devised.
As no large datasets with holes exist, we instead choose to use a reinforcement learning approach in which our reward function encourages the model to make ``long'' predictions with as few ``\hole'' tokens as possible, but to avoid making incorrect predictions.
We found that standard left-to-right sequence models perform poorly on this task.
Hence, we developed \projName, a model that construct suggestions by generating a (partial) syntax tree, but which has the option of leaving non-terminals in its output.

\begin{figure}
    \centering
\begin{minipage}[c]{0.29\columnwidth}
\textbf{Code Context}:
\vspace*{-1.5ex}
\begin{lstlisting}[numbers=left]
import sys
target = sys.argv[1]
(*\large \color{blue} I*)
\end{lstlisting}
\vspace*{-1ex}
\textbf{Ground-Truth}:
\vspace*{-1.5ex}
\begin{lstlisting}
ID = sys.argv[2]
\end{lstlisting}
\vspace*{-1ex}
\end{minipage}
\begin{minipage}[c]{0.7\columnwidth}
\textbf{Suggested Code Completions}:\\
\begin{tabular}{ll}
     \lrbaseline      & \texttt{target = target.replace("\textbackslash\textbackslash", "/")} \\
     \lrstopbaseline  & \texttt{target = }\\
     \lrholebaseline  & \texttt{print(target)}\\ 
     \texttt{Copilot} & \emph{(No suggestion)}\\
     \projName        & \texttt{\hole = sys.argv[2]}
\end{tabular}
\end{minipage}
    \caption{A sample snippet (left; abbreviated from \autoref{fig:python-outputs-3} in \autoref{appx:generated samples}).
    A developer has just typed the code and their cursor (in {\color{blue}blue}) is at line 3.
    Code completions provided by a number of models are shown on the right, where \lrbaseline is a standard \lmc and \projName is our new model.}
    \label{fig:qual example}
\end{figure}

\textbf{Contributions}
(1) We present \projName, a transformer-based model that generates code based on the programming language grammar and can predict hole tokens rather than output it is uncertain about.
(2) We develop \regexAcc, a metric that can evaluate the fit of predictions with holes.
(3) We evaluate \projName on large corpora of Python and C\# code and show that \projName can make longer and more precise statement-level sketch completions compared to a number of baselines.

\section{Method}
\label{sec:method}
Our aim is to predict code completions as \emph{sketches}, a mix of actual tokens and ``holes'' \hole, which are meant to signify that the model is unable to make a useful prediction within the given context and further user input is required.
Formally, we consider models that take a context sequence $\inSeq$ of tokens as input and have to produce an output sequence $\outSeq$; intuitively, $\inSeq$ is what the user typed so far, and $\outSeq$ is the suggestion presented to the user.
In our setting, $\outSeq$ is a \emph{sketch}, a mix of tokens from the programming language and the special token $\hole$ signifying a ``hole'' that could be filled by an arbitrary sequence of tokens.
For example, \texttt{t = foo(\hole)} is a sketch corresponding to assigning the return value of function \texttt{foo} to variable \texttt{t}, but leaves the arguments of the function call undetermined.

\paragraph{Metric}
A good sketch is one that
 (a) can be completed into the correct output and 
 (b) is as precise as possible.
To measure how successful a method is in doing so, we define a new metric \regexAcc.
For (a), we use $\texttt{toRegex}(\pred{\outSeq})$ to turn a predicted code sketch $\pred{\outSeq}$ into a regular expression by replacing all holes with the wildcard matching any non-empty sequence (``\texttt{.+}'' in Perl Compatible Regular Expression syntax).
If the regex matches the ground truth, \texttt{matches($\cdot, \cdot$)} returns a score of 1 otherwise it returns 0.
To implement (b), we scale this result by the proportion of terminal tokens predicted, by defining $\texttt{nTokens}(\pred{\outSeq})$ as the function that returns the number of non-hole symbols in $\pred{\outSeq}$.
More formally, assume an output sketch $\pred{\outSeq}$ and a ground-truth sequence $\gt{\outSeq}$, where $\gt{\outSeq}$ does \emph{not} contain any \hole tokens.
\regexAcc is then defined as
\begin{align*}
    \text{\regexAcc}(\pred{\outSeq}, \gt{\outSeq}) \triangleq
    \texttt{matches}(\texttt{toRegex}(\pred{\outSeq}), \gt{\outSeq})
    \cdot \frac{\texttt{nTokens}(\pred{\outSeq})}{\texttt{nTokens}(\gt{\outSeq})}.
\end{align*}

Beyond \regexAcc, we also consider ROUGE~\citep{lin2004rouge}, since a sketch can be thought as a form of a ``summary'' of the target text.
For this, we use a helper function \textsc{EraseHoles}($\pred{\outSeq}$) that simply drops all $\hole$ tokens, and then consider $\rouge_{\text{F1}}(\textsc{EraseHoles}(\pred{\outSeq}), \gt{\outSeq})$.
\rouge is more lenient to errors than \regexAcc and gives partial credit to non-matching but plausible sketches.

\subsection{Linear Code Sketch Generation}
First, we consider the idea of generating code sketches using a standard generative model for language.
To this end, we simply extend the vocabulary with the special ``$\hole$'' token.
An obvious problem is that while we have plenty of training data for a standard generative model, we do not have training data for outputs $\outSeq$ that contain the \hole token.
Consequently, we cannot train the model in a fully supervised fashion, and instead turn to reinforcement learning.
Concretely, we devise a reward function $r(\cdot)$ that averages \regexAcc and \rouge, \ie for a predicted output sketch $\pred{\outSeq}$ and a ground truth output (without \hole tokens) $\gt{\outSeq}$, we define
\begin{align}\label{eq:reward fn}
    r(\pred{\outSeq}, \gt{\outSeq}) = 
        \frac{1}{2}
            \left(
                  \regexAcc(\pred{\outSeq}, \gt{\outSeq}) 
                + \rouge_{\text{F1}}(\textsc{EraseHoles}(\pred{\outSeq}, \gt{\outSeq})
            \right).
\end{align}
Using the combination of \rouge (which does not consider holes) and \regexAcc is crucial here, as \rouge is much ``smoother'' compared to \regexAcc, which is 0 for all but very few predictions, allowing us to measure partial improvement.
We use our reward function from \autoref{eq:reward fn} to evaluate the quality of the output of the full model and compute a loss.
Inspired by \citet{paulus2017deep} we use self-critical policy
gradient training~\citep{rennie2017self} and for a prediction $\pred{\outSeq}$ we minimise
\begin{align}
    \label{eq:seq model loss}
    \mathcal{L}(\inSeq, \gt{\outSeq}) =
        \left(r(\pred{\outSeq}, \gt{\outSeq}) - \tilde{r}(\inSeq) \right)
        \cdot
        \mathcal{L}_{\text{gen}}\left(\inSeq, \pred{\outSeq}\right)
\end{align}
Here, $\tilde{r}(\inSeq)$ is the reward achieved by the prediction from the snapshots of the model that achieved the best score so far and $\mathcal{L}_{\text{gen}}$ is the loss of the generative model. Intuitively, this objective rewards models that improve upon the previous best policy with respect to $r$.

To model this in practice, we use a standard encoder/decoder Transformer model~\cite{vaswani2017attention,radford2019language},
``translating'' the context $\inSeq$ into the output $\outSeq$ using separate encoder and decoder models.
We additionally also consider the language modelling case, \ie, a model that conditioned on $\inSeq$ predicts token $\outSeq_{0}$, conditioned on $\inSeq, \outSeq_{0}$ predicts token $\outSeq_{1}$, \etc.

\paragraph{Pretraining}
In practice, we found that directly training a sequence model to maximise \autoref{eq:reward fn}, is very slow and does \emph{not} converge to a useful model.
Instead, we heuristically generate a dataset suitable for supervised pretraining. We replace random AST non-terminals of the target output by \hole and generate target sequences. These contain terminals and zero or more \hole.
We then pretrain the model on this dataset to convergence, and then fine-tune it using the reward of \autoref{eq:reward fn}.

\begin{figure}[t]
    \newcommand{\term}[1]{{\color{blue}\code{#1}}}
\newcommand{\selPos}[2]{{\color{gray}\tiny$i^{(#1)}=#2$}}
\newcommand{\inputSeq}[1]{{\color{gray}\tiny$\mathbf{x}^{(#1)}$:}}
\begin{tabular}{rllllllllllll}
\inputSeq{0} & \term{r =} & \multicolumn{10}{c}{\NT{Expr}}                                                                                                         & \selPos{0}{3} \\ \cline{3-12}
\inputSeq{1} & \term{r =} & \NT{Expr} & \term{*}  & \multicolumn{8}{c}{\NT{ParenthesizedExpr}}                                                                     & \selPos{1}{5}\\ \cline{5-12}
\inputSeq{2} & \term{r =} & \NT{Expr} & \term{*}  & \term{(} & \multicolumn{6}{c}{\NT{Expr}}                                                            & \term{)} & \selPos{2}{6}\\ \cline{6-11}
\inputSeq{3} & \term{r =} & \NT{Expr} & \term{*}  & \term{(} & \NT{Expr}          & \term{-} & \multicolumn{4}{c}{\NT{Expr}}                             & \term{)}& \selPos{3}{8}\\ \cline{8-11}
\inputSeq{4} & \term{r =} & \NT{Expr} & \term{*}  & \term{(} & \NT{Expr}          & \term{-} & \NT{Identifier} & \term{(} & \NT{ArgList} & \term{)} & \term{)}     & \selPos{4}{8}\\ \cline{8-8}
\inputSeq{5} & \term{r =} & \NT{Expr} & \term{*}  & \term{(} & \NT{Expr}          & \term{-} & \term{foo}      & \term{(} & \NT{ArgList} & \term{)} & \term{)}     & \selPos{5}{10}\\ \cline{10-10}
\inputSeq{6} & \term{r =} & \NT{Expr} & \term{*}  & \term{(} & \NT{Expr}          & \term{-} & \term{foo}      & \term{(} & \NT{Identifer}    & \term{)} & \term{)}& \selPos{6}{10} \\ \cline{10-10}
\inputSeq{7} & \term{r =} & \NT{Expr} & \term{*}  & \term{(} & \NT{Expr}          & \term{-} & \term{foo}      & \term{(} & \term{args}       & \term{)} & \term{)}& \selPos{7}{6}\\ \cline{3-3}
\inputSeq{8} & \term{r =} & \NT{Identifier} & \term{*}  & \term{(} & \NT{Expr}    & \term{-} & \term{foo}      & \term{(} & \term{args}       & \term{)} & \term{)}& \selPos{8}{6}\\ \cline{3-3}
\inputSeq{9} & \term{r =} & \term{x} & \term{*}  & \term{(} & \NT{Expr}        & \term{-} & \term{foo}      & \term{(} & \term{args}       & \term{)} & \term{)}& \selPos{9}{\stopExpand}\\
\end{tabular}
    \caption{Progress of grammar-based code generation of the sketch \code{r = x * (\hole - foo(args))} by \projName.
    Each line represents consecutive $\inSeq^{(t)}$ in \autoref{alg:generation}.
    Terminal tokens are shown in \term{monospace blue} font. 
    The underlined non-terminal at position $i^{(t)}$ is selected by \selectorModel and its expansion is generated by \expansionModel, \ie the output underneath the selected (underlined) non-terminal.
    \autoref{fig:charp-generation-2} and \autoref{fig:python-generation-1} in \autoref{appx:generated samples} show real example generation sequences from our datasets.}\label{fig:generation example}
\end{figure}

\subsection{Grammar-Based Code Sketch Generation}
In our experiments, we found the simple extended sequence model from above to not perform well, in particular, \hole tokens would not replace semantically meaningful subsequences (\eg  ``\texttt{szconv.\hole)}'' does not contain a left parenthesis and requires the user to fill it in.).
To resolve this, we developed \projName, a \emph{grammar-guided} model.
It generates code by following the structure of the context-free grammar (CFG) defining the programming language syntax, iteratively expanding non-terminal symbols.
Crucially, it can choose to \emph{not} expand some non-terminal symbols, which can then be presented as \hole to the user.
In traditional grammar-based generation of text~\citep{cohen2012spectral} or code~\citep{maddison2014structured,yin2017syntactic,allamanis2014mining,bielik2016phog}, the CFG is followed by sequentially expanding the left-most, bottom-most non-terminal symbol, using one of the production rules of the grammar.
\projName changes this and instead selects which (if any) non-terminal symbol to expand.
An example generation is displayed in \autoref{fig:generation example}.

\paragraph{Probabilistic Model}
A CFG is defined as a tuple $(\Sigma, \mathcal{N}, S, R)$ where $\Sigma$ is a set of terminal symbols, $\mathcal{N}$ is a set of non-terminal symbols, $S\in \mathcal{N}$ is the root symbol and $R$ is a set of production rules.
We denote non-terminals as \NT{NonTerminalName}.
\projName can be viewed as a sequence-to-sequence model transforming $\inSeq = \inSeq_{0}, \inSeq_{1}, ..., \inSeq_{n}$ into a new sequence in which one non-terminal symbol $x_i$ has been replaced by a new sequence of new symbols, according to a production rule of the grammar.
Examples of such sequences and rewrites are shown in \autoref{fig:generation example}.

\projName does this rewriting in two steps.
First, a \emph{non-terminal selector model} \selectorModel selects a non-terminal in \inSeq to expand and then the \emph{non-terminal expansion model} \expansionModel determines how to expand it.
To define \selectorModel, let $N(\inSeq) = \{i \mid \inSeq_{i} \in \mathcal{N}\} \cup \{ \stopExpand \}$ denote the set of non-terminal positions in \inSeq and a special ``stop expansion'' \stopExpand symbol.
Conditioned on \inSeq, \selectorModel produces a probability distribution over $N(\inSeq)$.
In turn, \expansionModel is conditioned on \inSeq and a position $i \in N(\inSeq)$ and models a probability distribution over expansion sequences $\expSeq \in (\Sigma \cup \mathcal{N})^*$.
Note that factorising \projName into two models \selectorModel and \expansionModel is an important modelling decision: how to best expand a non-terminal is entirely separated from predicting whether a hole should be introduced.
These two concepts are intermixed in standard (sequence) decoders.
In practice, we define both models using neural architectures with partially shared parameters, as discussed below.

\begin{algorithm}[t]
    \caption{\projName generative process, given an input sequence $\inSeq^{(0)}$.}\label{alg:generation}
\begin{algorithmic}
\For{$t=0,1,2,...$} 
    \State{%
        $
        i^{(t)}
        \sim
        \selectorModel\left(
            i
            \mid
            \inSeq^{(t)}, N(\inSeq^{(t)})
        \right)
        $
    }\Comment{sample non-terminal position from $N(\inSeq^{(t)})$ to expand}
    \If{$i^{(t)} = \stopExpand$} \Comment{if $\inSeq^{(t)}$ does \emph{not} contain non-terminals or none was selected by \selectorModel}
        \State\textbf{break}\Comment{stop generation}
    \EndIf
    \State{%
        $
        \expSeq_{\circledcirc i^{(t)}}^{(t)}
        \sim
        \expansionModel\left(
            \expSeq
            \mid
            \inSeq^{(t)}, i^{(t)}
        \right)
        $
    }\Comment{sample expansion of non-terminal at position $i^{(t)}$}
    \State{%
        $
        \inSeq^{(t+1)}
        \leftarrow
        \inSeq_{<i^{(t)}}^{(t)}
        :: \expSeq_{\circledcirc i^{(t)}}^{(t)}
        :: \inSeq_{>i^{(t)}}^{(t)}
        $
    }\Comment{create $\inSeq^{(t+1)}$ by replacing non-terminal at $i^{(t)}$ by $\expSeq_{\circledcirc i^{(t)}}^{(t)}$}
\EndFor
\Return $\textsc{NonTerminalsToHoles}(\inSeq^{(t)})$\Comment{convert remaining non-terminals to holes and return}
\end{algorithmic}
\end{algorithm}

\autoref{alg:generation} shows a high-level description of \projName, in which \selectorModel and \expansionModel are used repeatedly to select and expand non-terminals (not necessarily the left-most one), until none are left or the \selectorModel indicates that expansion should stop.
Here, $\textsc{NonTerminalsToHoles}(\cdot)$ replaces all remaining non-terminal symbols with a hole \hole.
Note that \projName is \emph{not} context-free, taking into account the whole input sequence when expanding a non-terminal.
Second, in contrast to many grammar-based methods~\citep{yin2017syntactic,bielik2016phog}, any non-terminal can be expanded at each step.
Finally, \expansionModel is not directly constrained to follow the production rule set $R$, but can generate any sequence.
In practice, it learns to follow to the rules of $R$ from the data, but this flexibility is important for handling string literals and argument tuples of variable length.

\paragraph{Neural Model}
To implement \selectorModel and \expansionModel, we use a shared encoder module that computes a representation of the input sequence $\inSeq = \inSeq_{0}, \ldots, \inSeq_{n}$ as vectors $\mathbf{e}_0, \ldots, \mathbf{e}_n$, $\mathbf{e}_i \in \mathbb{R}^{D}$, where $D$ is a hyperparameter.
Our encoder module is a Transformer~\citep{vaswani2017attention}, given the impressive results of transformer-based models in NLP and code~\citep{feng2020codebert}.
Other architectures (RNNs, 1D-CNNs, Transformer variants) would be suitable as well, but we leave their study for future work.

\selectorModel is implemented similar to a pointer network on top of this encoder module, i.e.
\begin{align*}
    \selectorModel(i \mid \inSeq) = \softmax_{i \in N(\inSeq)} \left( f(\mathbf{e}_i) \right),
\end{align*}
where $f$ is a learnable feed-forward neural network.
For our purposes, we define $\mathbf{e}_\stopExpand$ as the representation of the special start symbol \texttt{[CLS]} used in our Transformer encoder.

The expansion model \expansionModel follows a standard autoregressive decoder formulation, \ie
\begin{align*}
    P_e(\expSeq \mid \inSeq, i) = \prod_{j=1}^m P_{dec}(\expSeq_{j} \mid \mathbf{e}_0, \ldots, \mathbf{e}_n, i, \expSeq_{<j}).
\end{align*}
We implement $P_{dec}$ as a (causal) relational Transformer decoder, similar to \citet{wang2019rat}.
Relational transformers augment the attention mechanism by incorporating predefined relationships among elements; attention scores are then biased by learnable weights for each relation.
In \projName, we only use a single relation, connecting each token to the expanded non-terminal token $\inSeq_{i}$, to help the model focus on the token it needs to generate an expansion for.

\paragraph{Objective}
Due to the lack of supervised data, we employ reinforcement learning to train \projName.
We use our reward function from \autoref{eq:reward fn} to evaluate the quality of the output of the full model.
We use self-critical policy gradient training as in \autoref{eq:seq model loss} and minimise
\begin{align}
    \label{eq:full loss}
    \mathcal{L}(\inSeq, \gt{\outSeq}) =
        \left(r(\pred{\outSeq}, \gt{\outSeq}) - \tilde{r}(\inSeq) \right)
        \cdot
        \sum_{t=0}^T 
            \left(
                - \log \selectorModel\left(i^{(t)} \mid \inSeq^{(t)}\right) 
                - \mathbb{I}\left(i^{(t)} \neq \stopExpand\right) 
                    \cdot
                    \log \expansionModel\left(\gt{(\expSeq_{\circledcirc i^{(t)}}^{(t)})} \mid \inSeq^{(t)}, i^{(t)}\right)
            \right).
\end{align}
Here, $\tilde{r}(\inSeq)$ is the reward achieved by the snapshots of \selectorModel and \expansionModel that achieved the best score so far.
The rest of the objective follows the iterations of the loop in \autoref{alg:generation}, where
    $t$ is the iteration index,
    $\pred{\outSeq}$ is the predicted sketch,
    $\gt{\outSeq}$ is the ground-truth sequence of terminals, and
    $\mathbb{I}(\cdot)$ is the indicator function.

\paragraph{Pretraining}
As in the sequence model, directly training with the RL objective \autoref{eq:full loss} is computationally intensive due to the sampling requirement.
We again use a pretraining strategy.
First, we train \expansionModel to expand every non-terminal, independently of the expansion order learned by \selectorModel.
To do this, we use the input training examples and follow \autoref{alg:generation}, but instead of sampling from $\selectorModel(\cdot)$, we sample $i^{(t)}$ from a uniform distribution over the non-terminals in $\inSeq^{(t)}$, $\widetilde{N}(\inSeq^{(t)}) = \{i \mid \inSeq_{i} \in \mathcal{N}\}$.
This yields sequences of intermediate sketches $\inSeq^{(t)}$ for each example.
Furthermore, for each $\inSeq^{(t)}$, we compute the ground-truth expansion $\gt{(\expSeq_{\circledcirc i}^{(t)})}$ for all non-terminals $i \in \widetilde{N}(\inSeq^{(t)})$.
We can then pretrain \expansionModel using the supervised objective
\begin{align*}
\mathcal{L}_{\text{pre, e}}\left(
    \inSeq^{(t)},
    \gt{(\expSeq_{\circledcirc i}^{(t)})}_{i \in \widetilde{N}(\inSeq^{(t)})}
\right)
= 
\frac{1}{\vert \widetilde{N}(\inSeq^{(t)}) \vert}
\cdot
\sum_{i \in \widetilde{N}(\inSeq^{(t)})}
    -\log \expansionModel\left(\gt{(\expSeq_{\circledcirc i^{(t)}}^{(t)})} \mid \inSeq^{(t)}, i\right),
\end{align*}
\ie the negative log-likelihood of the correct expansion for \emph{all} non-terminals in $\inSeq^{(t)}$.
This computation is more computationally efficient compared to the one in \autoref{eq:full loss} since the cost of encoding $\inSeq^{(t)}$ is amortised across all potential expansions and no sampling is required.
Once \expansionModel is pretrained, we pretrain \selectorModel.
For this, we fix the weights of the shared encoder module, and optimise only the remaining parameters of \selectorModel through \autoref{eq:full loss}.
Once we have a pretrained both models, we then fine-tune all model weights end-to-end, using \autoref{eq:full loss}.

\paragraph{Optimisation: Grammar Flattening}
Following the formal grammar of a programming language commonly introduces tedious expansions.
For example, the Python non-terminal \NT{Call} is always expanded to \code{\NT{Expr}(\NT{ArgumentList})}, and the C\# non-terminal \NT{NotEqualOp} is always expanded to the terminal \code{!=}.
We ``flatten'' the grammar by replacing non-terminals such as \NT{Call} and \NT{NotEqualOp} with all their possible expansions.
In \autoref{appx:expansions} we provide the list of the flattened non-terminals.
Note that if we repeated this process for all non-terminals except from the starting symbol $S$, \projName would degenerate into a standard encoder-decoder model.

\paragraph{Beam Search}
At test time, we employ a two-step beam search, and replace sampling from \selectorModel and \expansionModel with their top-$\nu$ outputs, keeping a beam of size $k$.
First, for each $\inSeq^{(t)}$ in the beam, we compute \selectorModel and select the top-$m$ non-terminal positions to expand.
For each of those $m$ positions, we sample the top-$n$ expansions from \expansionModel using a standard beam search.
We compute the likelihood of all $k \cdot n \cdot m$ results, and then keep only the top-$k$.
This process (detailed in \autoref{appx:beam search}) is similar to a standard beam search but takes into account that two submodels are used.

\paragraph{Computational Cost}
\projName's ability to predict sketches comes with additional computational cost compared to standard transformer encoder-decoder models: since at each iteration of the loop in \autoref{alg:generation} $\inSeq^{(t)}$ changes, \selectorModel and \expansionModel must be recomputed.
This means that the encoder-decoder needs to run once on each partial sequence, in contrast to left-to-right causal generation, in which intermediate results can be re-used.
Future work may consider selecting more than one element to expand from $N(\inSeq^{(t)})$ to expand at each step, reducing the total number of expansion steps, similar to \citet{welleck2019non,stern2019insertion}.

\section{Evaluation}
\label{sec:evaluation}
To empirically evaluate the ability of our models to learn to predict useful code completions, we use the metrics \regexAcc and \rouge as discussed above.
Note that we measure these only on the newly generated output sequence, \ie we ignore the ``prompt'' of context tokens.

\paragraph{Datasets}
To collect a dataset, we clone all non-fork repositories with more than 20 stars on GitHub that have C\# or Python as their top language.
Then, we deduplicate the corpus using the method of \citet{allamanis2019adverse,lopes2017dejavu}.
Finally, we parse all files into a syntax tree using \href{http://tree-sitter.github.io/tree-sitter/}{Tree-sitter},
ignoring any files that cannot be parsed using the v0.19.0 grammar definitions.
Finally, we split the files into 70-10-20 train-validation-test.
To create (pre-)training examples, \ie inputs to \autoref{alg:generation}, we search the syntax tree of each file and for each \NT{SimpleStatement} non-terminal create an example.
The syntax tree rooted at the \NT{SimpleStatement} non-terminal is then used to get the ground-truth expansions during pre-training and the ground-truth expansion \gt{\outSeq}.
For our test set, we randomly sample a \NT{SimpleStatement} non-terminal for each file to evaluate and obtain 318K (resp. 362K) examples for C\# (resp. Python).
For each example, \inSeq is the 200 terminal tokens \emph{before} the \NT{SimpleStatement} non-terminal.
More details about the dataset can be found in \autoref{appx:dataset stats}.

\paragraph{Baselines}
Since we are not aware of any prior model that targets code completion with sketches, we consider two Transformer-based baselines.
We consider both the sequence-to-sequence setting using separate encoder and decoder models~\citep{vaswani2017attention} as well as the language modelling setting (where there is no distinction between encoder and decoder).
We refer to these as ``\lrbaseline'' and ``\lmbaseline''.
We use ``\lrbaseline'' to denote a standard Transformer encoder-decoder model~\citep{vaswani2017attention} used in sequence-to-sequence tasks.
Additionally, we consider ``\lrstopbaseline'' and ``\lmstopbaseline'', which are trained to stop generation by inserting a final \hole token that captures any suffix.
Note that this models can only generate sketches that are prefixes of the target completion, \ie it corresponds to a standard token-level generative model with a learnable stopping ability.
To train this model, we use self-critical policy gradient training (as in \autoref{eq:seq model loss}).

\paragraph{Model Training}
We provide the training details for all experiments.
Most of our models use a 6-layer Transformer as encoder and 6-layer Transformer as decoder, each with a hidden dimension of 768 and 12 attention heads, with the exception of the \lmbaseline model (and its variations), which uses a single 12-layer Transformer, to match the number of parameters of the other models.
We set the intermediate dimension of each Transformer layer as 3072 and use 3 fully-connected layers with 3072, 768 and 1 hidden sizes as the feed-forward neural network $f$ in the selector model \selectorModel.
The vocabulary is constructed using byte-pair encoding \citep{sennrich2015neural} and the vocabulary size is $25\,000$.
We set max length of input and output sequences as 512 and 64, respectively.
We train the model with Adam optimiser using a learning rate of 2e-5 and batch size $4\,096$. We used automatic mix precision.
Training was performed on 64 NVIDIA Tesla P100 with 16GB memory for 10 days.
During beam search we use $k=5, n=1$ and $m=\infty$, \ie we consider all non-terminal positions in each $\inSeq^{(t)}$.
We selected these numbers during early experiments as a reasonable trade-off between speed and predictive performance.

\paragraph{Results}

\begin{table}[tb]\centering
\small
\caption{Performance of \projName compared to baselines for Python and C\#.}\label{tbl:main results}
\begin{tabular}{@{}lrrrrrrrrr@{}}
  \toprule
                                 &                    \multicolumn{4}{c}{C\#}                   &&              \multicolumn{4}{c}{Python} \\
                                                \cmidrule{2-5}                                                  \cmidrule{7-10}
                                 &  \multicolumn{2}{c}{\regexAcc}  &    \rouge    & Avg  && \multicolumn{2}{c}{\regexAcc}&     \rouge   &  Avg \\
                                                \cmidrule{2-3}                                                \cmidrule{7-8}
                                 &    Top 1      &           Top 5 &              &    Len      &&     Top 1    &      Top 5    &              & Len \\
  \midrule
    \lmbaseline                  &          0.42 &            0.52 &         75.7 &\textbf{8.0} &&         0.18 &          0.24 &         51.0 & \textbf{8.6}\\
    \lrbaseline                  &          0.42 &            0.47 &         77.0 &         7.1 &&         0.17 &          0.20 & \textbf{53.2}&         5.8 \\
    \lmstopbaseline              &          0.42 &            0.49 &         70.9 &         6.8 &&         0.19 &          0.25 &         49.5 &         7.3 \\
    \lrstopbaseline              &          0.45 &            0.54 &         69.1 &         5.3 &&         0.20 &          0.29 &         39.3 &         3.0 \\
    \lmholebaseline              &          0.44 &            0.54 &         73.3 &         6.3 &&         0.20 &          0.27 &         53.9 &         6.6 \\
    \lrholebaseline              &          0.45 &            0.55 &         73.5 &         5.8 &&         0.18 &          0.22 &         48.9 &         4.7 \\
    \projName (pre-trained only) &          0.45 &            0.57 &         77.0 &         7.2 &&         0.20 &          0.29 &         50.2 &         5.7 \\    
    \projName                    &  \textbf{0.47}&    \textbf{0.59}& \textbf{77.4}&         7.5 && \textbf{0.21}&  \textbf{0.30}&         51.6 &         6.1 \\
  \bottomrule
\end{tabular}
\end{table}

\autoref{tbl:main results} shows the results for all considered models.
For both Python and C\#, \projName outperforms the baseline methods in terms of \regexAcc,
showing that the grammar-based generation can create better sketches
compared to simpler methods.
Note that although \lrbaseline has a 
comparable or better \rouge score, it does substantially worse than \projName with respect to
\regexAcc, meaning that the predictions are ``similar''
but the sketches contain errors (\ie do not match the ground-truth). This means that if a code
completion system suggested the full output of \lrbaseline,
the user would have to pause and correct the suggestion more frequently.
On the other hand, \lrstopbaseline improves over \lrbaseline in terms of \regexAcc
but has a worse \rouge and generates significantly shorter
suggestions (5.3 \vs 7.5 tokens-long for C\#). This is
expected since \lrstopbaseline is trained to be more
``conservative'' (\ie avoid incorrect suggestions) but
is also unable to introduce holes beyond the last
generated token.
Finally, we can see that while \projName already performs well after our pre-training procedure, we can further improve its performance with our fine-tuning technique.
We believe that this is because when \selectorModel and \expansionModel are trained jointly, they co-adapt: some of the capacity of the shared encoder module that is used to make predictions for hard-to-expand non-terminals is ``freed'' since \selectorModel learns to not expand them.

\begin{figure}[t]
\begin{subfigure}[b]{0.47\textwidth}
    \centering
    \includegraphics[width=\textwidth]{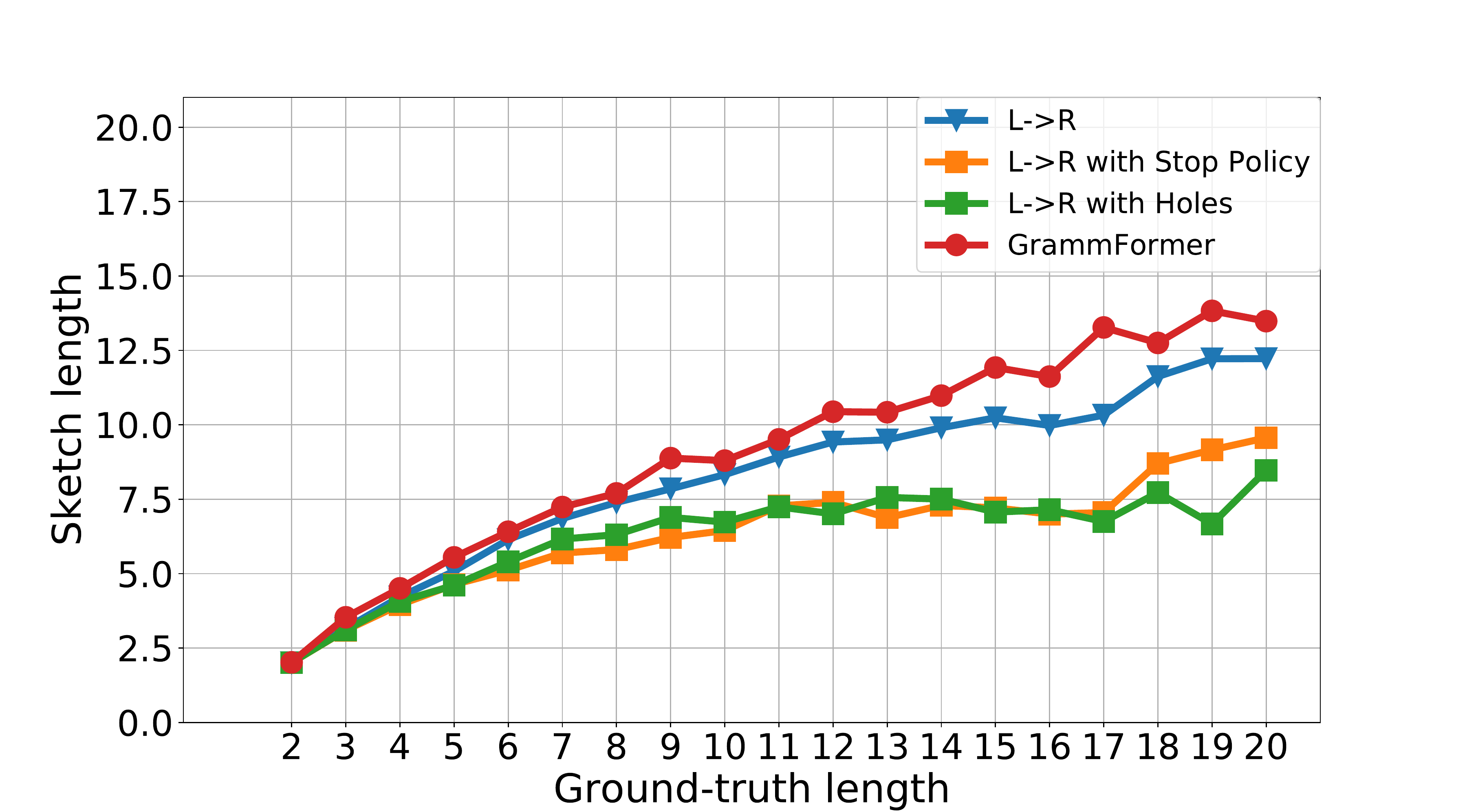}
    \caption{C\#}
    \label{fig:c}
\end{subfigure}
\hfill
\begin{subfigure}[b]{0.47\textwidth}
    \centering
    \includegraphics[width=\textwidth]{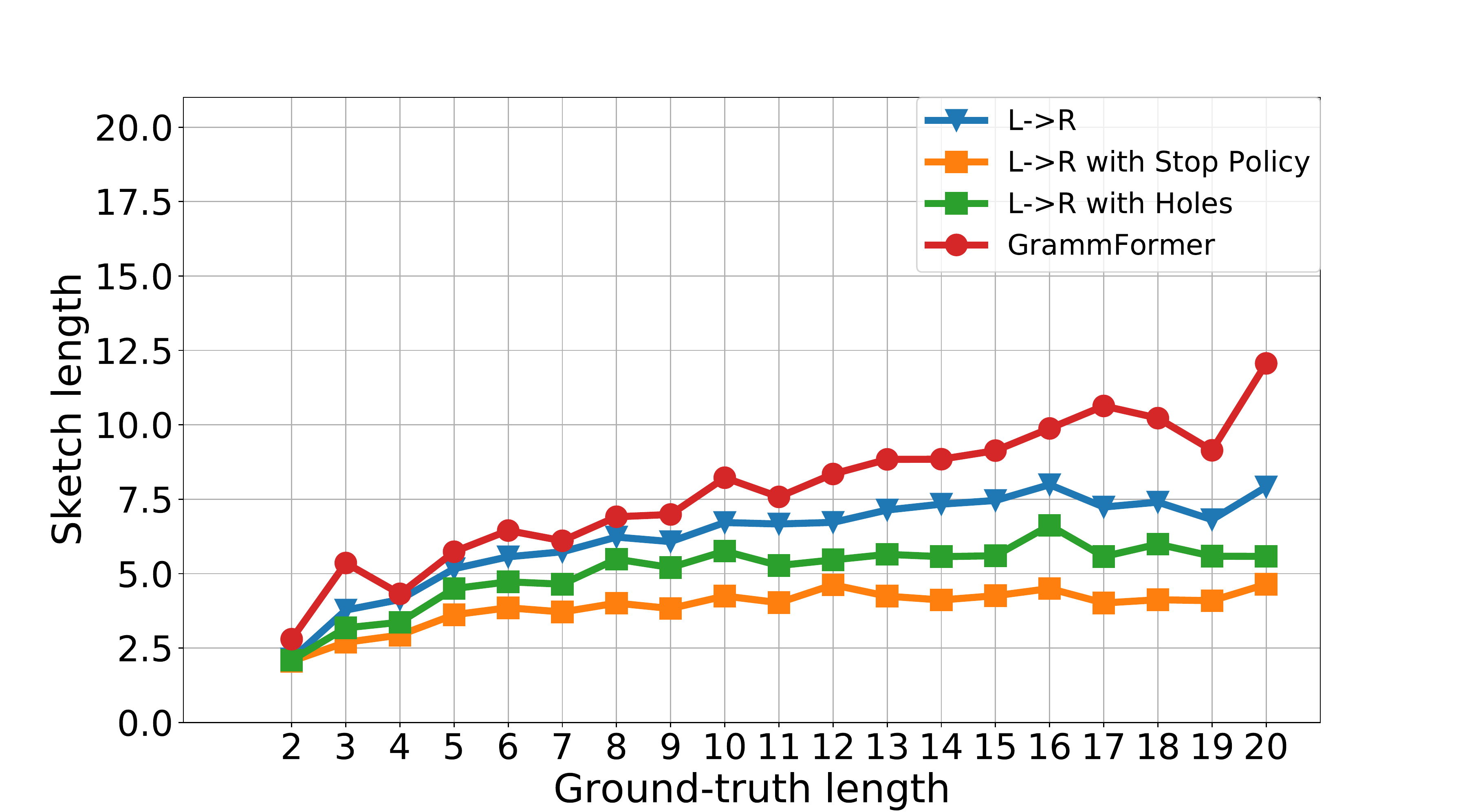}
    \caption{Python}
    \label{fig:d}
\end{subfigure}
\caption{Sketch length \vs ground-truth length}\label{fig:sketch len vs ground length}
\end{figure}

\autoref{fig:sketch len vs ground length} shows how the length of the generated code sketch relates to the length of the ground truth expression.
While the differences between models are small for short target sequences, \projName generates substantially longer suggestions than other models when more complex suggestions are required.
In particular, the \lrstopbaseline model generates very short suggestions, as it is trained to stop generation whenever it reaches a point at which it is uncertain about the next token.

\begin{figure}[t]
\begin{subfigure}[b]{0.47\textwidth}
    \centering
    \includegraphics[width=\textwidth]{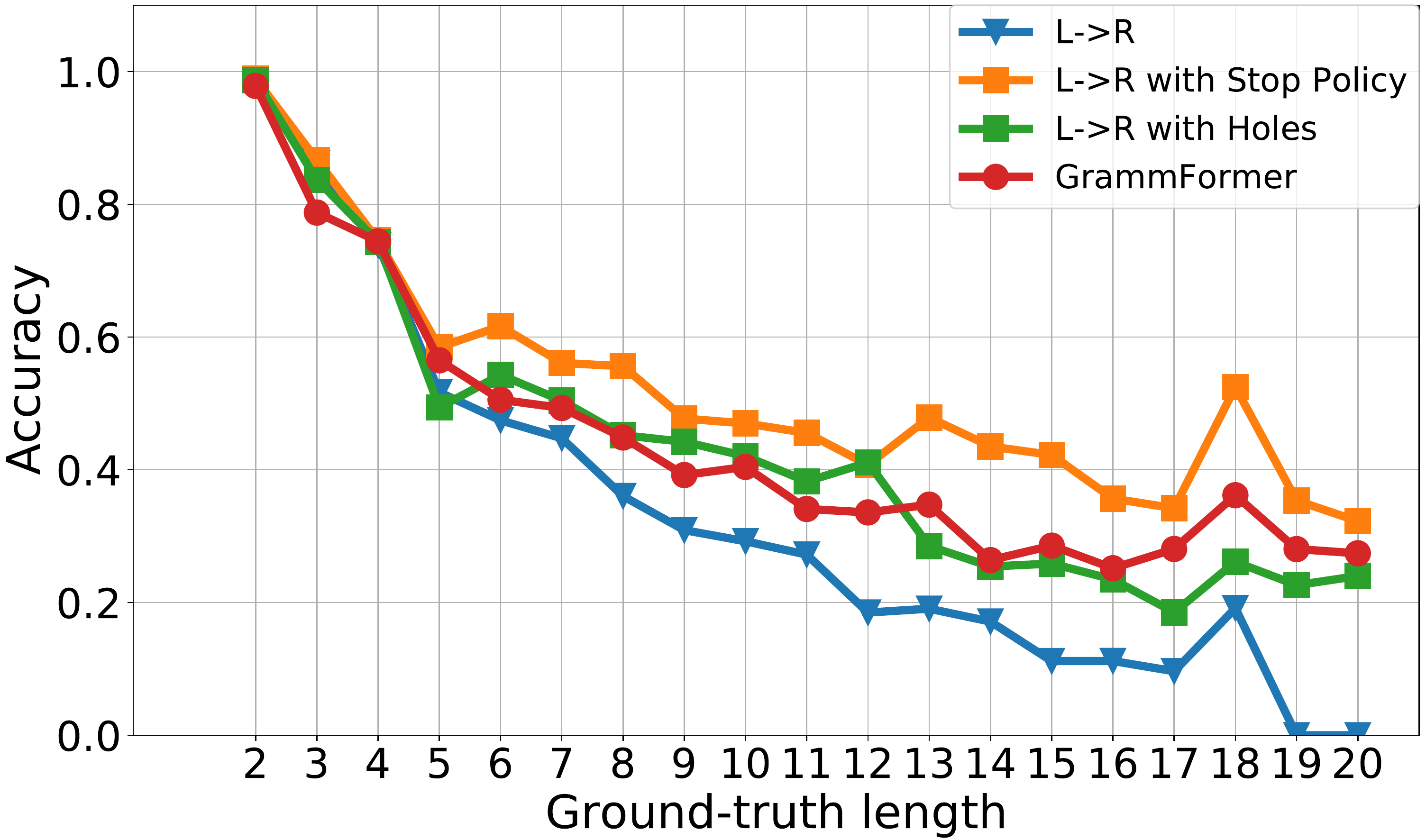}
    \caption{C\#}
    \label{fig:e}
\end{subfigure}
\hfill
\begin{subfigure}[b]{0.47\textwidth}
    \centering
    \includegraphics[width=\textwidth]{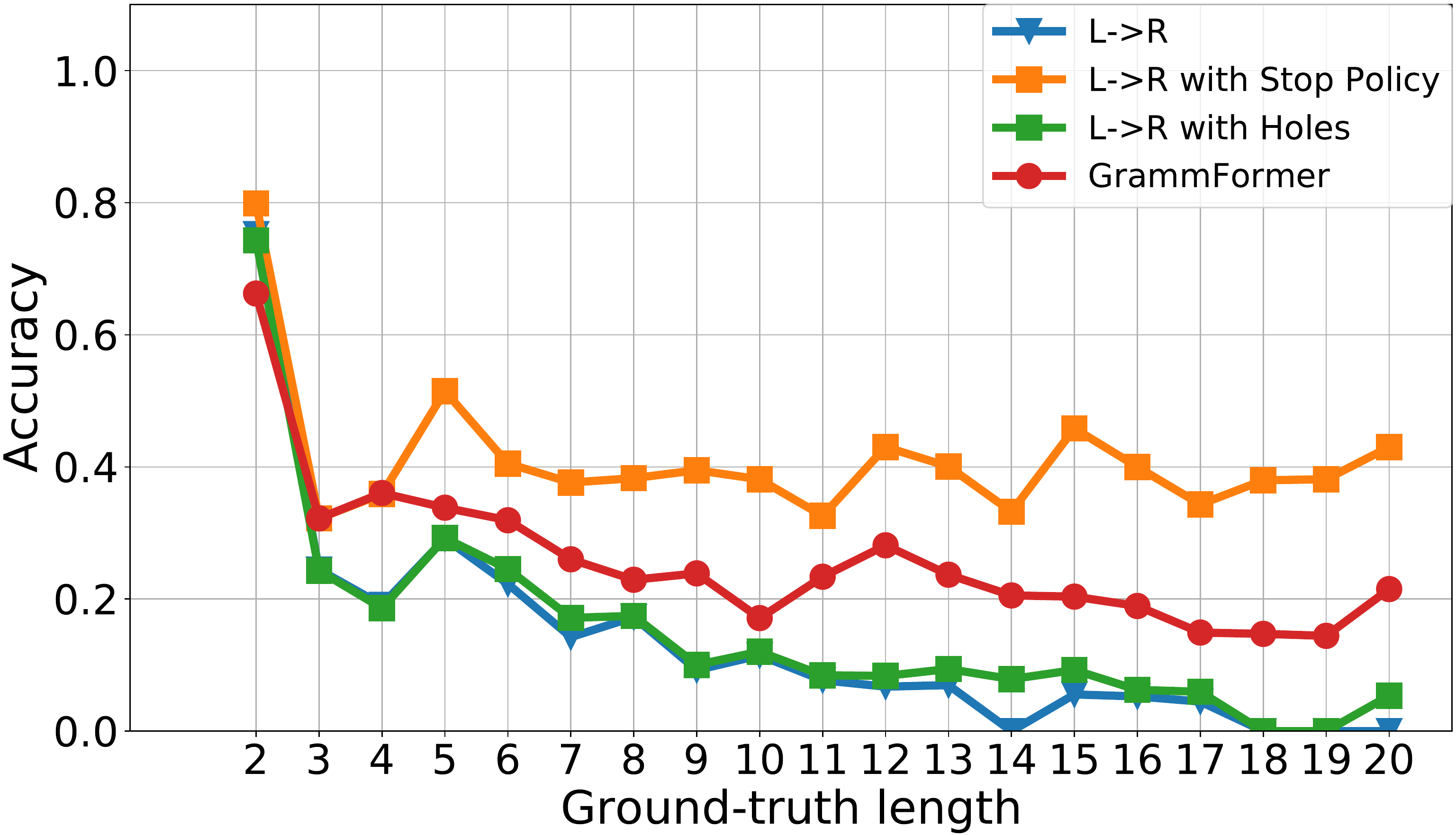}
    \caption{Python}
    \label{fig:f}
\end{subfigure}
\caption{Percent Correct (\ie, matching) sketches (top-1 generated sketch) \vs ground-truth length}\label{fig:accuracy len vs ground length}
\end{figure}

\autoref{fig:accuracy len vs ground length} in turn shows how often the suggested sketch was correct dependent on the length of the ground truth token sequence.
Here, \lrstopbaseline does best because it generates the shortest (\ie, least determined) predictions, which is exactly the trade-off captured by our \regexAcc metric.
Of the models that generate longer suggestions, \projName clearly does best, with the improvement becoming more pronounced with the length of the target sequence.
Note that the performance of the models on C\# is generally better compared
to the performance in Python. We believe that this has to do with the grammar
of each language and the patterns it induces within the developer's code. \citet{casalnuovo2019studying,karampatsis2019maybe} have observed a similar phenomenon on the
perplexity across (standard left-to-right) language models for different programming languages.

\begin{table}[tb]\centering
\caption{Performance for \projName ablations (C\#), for different \selectorModel and reward functions.}\label{tbl:policy}
    \begin{tabular}{@{}lrrrr@{}}
    \toprule
                                                             & \multicolumn{2}{c}{\regexAcc} & \rouge& Avg Length \\
                                                                     \cmidrule{2-3}
                                                             &       Top 1   &     Top 5     &       &  \\
    \midrule
\projName                                                    &          0.47 &     0.59      &  77.4 &  7.5 \\
    \midrule
Random expansion, no \stopExpand                             &          0.42 &     0.54      &  78.3 &  8.1 \\
Random expansion, \stopExpand at fixed threshold             &          0.45 &     0.57      &  71.6 &  5.8 \\
\midrule
\projName, $r(\cdot)=\rouge_{F1}$                            &          0.42 &     0.54      &  78.2 &  8.1 \\
\projName, $r(\cdot)=\regexAcc$                              &          0.51 &     0.62      &  70.8 &  5.8 \\
\midrule
\lrbaseline                                                  &          0.42 &     0.47      &  77.0 &  7.1 \\
\projName, no \stopExpand                                    &          0.42 &     0.55      &  78.1 &  8.2 \\
\bottomrule\end{tabular}
\end{table}

\paragraph{Ablations}
Next, we look into ablations of \projName and reason about how its components perform.
To this end, \autoref{tbl:policy} shows the performance of different model variants on the C\# dataset.

First, we analyse the effect of the selector model \selectorModel.
To this end, we consider two ablations.
The first is the ``random expansion'' model, in which the non-terminal token to expand is sampled uniformly at random from the full set of non-terminal symbols, and which hence does not stop expansion as long as any holes are remaining.
This is effectively \projName after our pre-training procedure for \expansionModel. 
This model achieves the best \rouge score, but a relatively bad \regexAcc, as it is forced to generate a prediction even when it is very uncertain.
The second ablation, a ``fixed threshold'' model is similar to our first ablation, but stops expansion when the probability of the generated $\inSeq^{(t)}$ falls below a threshold.
We choose this threshold on the validation set.
This model makes shorter, but more accurate sketch predictions compared to the ``random expansion'', but is worse than \projName.
These two ablations demonstrate that our learned \selectorModel is required for best performance.

Second, we consider the effect of using different reward functions $r(\cdot)$ in the training of \projName.
Concretely, whereas we use the mean of \regexAcc and \rouge in \projName, we now consider using only a single of these two metrics.
As expected, the results on the corresponding metric improve, with a substantial cost in the other metric.
Concretely, using only \regexAcc leads to significantly shorter predictions with a low \rouge score.
We believe that this is because \regexAcc is a strict metric, returning 0 if the sketch does \emph{not} match, which leads to sparse rewards and makes the resulting model more conservative at expanding non-terminals.

Finally, to evaluate the benefit of the grammar-guided decoder, we consider a variant of \projName that does not allow the introduction of \hole and instead has continue expansion until no non-terminals exist anymore.
This variant can be compared to \lrbaseline, which also cannot stop or introduce \hole tokens.
Our ablation shows that there is substantial benefit in using grammar-guided decoding, leading both to longer predictions as well as more correct ones.

\subsection{Qualitative Evaluation}

Having observed the quantitative results, we now turn our attention to
a qualitative look at the results and show some cherry-picked examples
that illustrate desired and undesired behaviours of \projName and the baselines, where we also include the suggestions of the GitHub Copilot system~\cite{copilot}.
 \autoref{fig:qual example} shows an example and eleven more are shown
in \autoref{appx:generated samples}.
\autoref{fig:qual example} illustrates the importance of generating sketches
instead of concrete sequences of terminal tokens: oftentimes, the code
context does not provide sufficient information about the user's intent.
Sketch-generating models can offer more informative suggestions given the partial intent.

Of course, \projName also makes mistakes. For example, \projName and \lrstopbaseline are sometimes ``too'' conservative (\eg \autoref{fig:csharp-bad-case-2} in \autoref{appx:generated samples}) generating holes where \lrbaseline generates fully concrete completions. This suggests future research opportunities for better calibration of $P_s$. 

Finally, a pure language modelling approach to code completion will always be insufficient.
For example, user-defined types and rare APIs cannot be predicted by a language
model, since the names of the APIs cannot be known during training (\autoref{fig:csharp-outputs-1} and \autoref{fig:python-bad-case-2} in \autoref{appx:generated samples}).
Researching methods to scalably introduce information from static analyses and additional context may alleviate this problem.

\section{Related Work}
One of the most successful applications of \lmcs is code completion
\citep{svyatkovskiy2019pythia,karampatsis2019maybe}. Transformer
LMs have been recently shown exceptional performance at 
the task being able to predict relatively long sequences of code~\citep{svyatkovskiy2020intellicode,chen2021evaluating}. Grammar-based code completion
and generation has been researched with neural \citep{maddison2014structured,yin2017syntactic,kim2021code}
and non-neural models \citep{bielik2016phog}, always expanding the left-most,
bottom-most non-terminal. In contrast to {\projName}s,
all these models target the generation of \emph{complete} code without
the ability to create sketches. R3NN~\citep{parisotto2016neuro}
generates only complete programs of a simple string transformation DSL
but expands the non-terminal with the highest confidence, instead of the left-most, bottom-most one, similar to \projName. In contrast to the aforementioned models,
\projName does \emph{not} maintain an explicit tree representation but
instead uses the sequences of leaves in the generation tree.

Sketch-like ideas appear in NLP such as the coarse-to-fine semantic parsing of \citet{dong2018coarse} and chat-bots of \citet{shum2019sketch}.
However, sketches are extracted deterministically to create a supervised dataset.
Similarly, SketchAdapt \citep{nye2019learning} uses a sequence model to generate sketches for small functional programs of a simple DSL towards speeding-up enumerative program synthesis from input-output examples. SketchAdapt is also trained as a supervised sketch generator. A supervised corpus is created by enumerating all possible sketches and selecting the one with the highest-probability and within a heuristically computed time budget. In \projName domain, enumerating all sketches is computationally intractable due to the complexity of general-purpose programming languages while no similar heuristic exists for code completion.

Recently, sequence generation approaches that go beyond the left-to-right
paradigm have been proposed~\citep{welleck2019non,stern2019insertion,gu2019levenshtein,ford2018importance,lee2018deterministic,shah2018generating},
usually by considering generation as an iterative refinement
procedure that changes or extends a sequence in every iteration.
These models often aim in speeding-up inference or allowing models to
figure a better order for generating a full sentence (of terminal tokens). However, since
these models focus on natural language and since its grammar is not defined a priori,
these methods do not follow a language grammar which effectively limits the space for sketch generation. Additionally, these work generate 
full utterances of text, rather than sketches. Future work may consider combining ideas in \projName with those models.

A related concept is learning to abstain~\citep{ziyin2019deep} where a model
learns to predict a ``don't know'' when it is uncertain about the outcome of a classification
task. This resembles the stop symbol ``\stopExpand'' with the difference that \projName
employs reinforcement learning to learn $P_s$ for a sequential problem rather than learning to abstain for a single-step classification problem.

\section{Discussion \& Conclusions}
In this work, we presented \projName, a generative model of
code that goes beyond standard left-to-right generation and 
is able to generate sketches, \ie snippets of code with holes.
Designing generative machine learning models with such abilities
is important towards facilitating better collaboration between
machine learning models and their human users.

While we have shown that \projName performs better than
other alternatives in the sketch generation task, there
are still many opportunities for improvement in the future.
First, larger transformer models will most probably yield better results,
as shown in the relevant literature.
Second, although we used \regexAcc as a plausible evaluation
metric, human studies for evaluating the trade-off between
sketch correctness and concreteness are needed. Such studies,
similar to those conducted for machine translation and summarization metrics,
can yield more informed reward functions $r(\cdot)$ and improved
user experiences.

Second, although we focused on programming languages,
modelling natural language also seems possible.
However, training such a model would require large corpora of
parsed text. 
Finally, we have treated programming languages as a sequence
of terminal and non-terminal symbols, ignoring the structure
imposed by code's strict semantics, such as data and control flow. Explicitly providing
information about the code's (deterministic) structure, \eg
with relational transformer encoders similar to \citet{hellendoorn2019global}
may further improve \projName's performance.

\subsubsection*{Acknowledgments}
The authors would like to thank Alex Polozov for useful discussions. We also thank
Patrick Fernandes, Szymon Malik, and Guilherme Ilunga for working on earlier modeling ideas on sketch generation. 
Although those were unsuccessful, they provided the inspiration for this work.

\bibliographystyle{iclr2022_conference}
\bibliography{bibliography}

\newpage
\appendix

\section{Generated Samples}
\label{appx:generated samples}
\begin{figure}[h]
	\begin{center}
		\includegraphics[width=1\columnwidth]{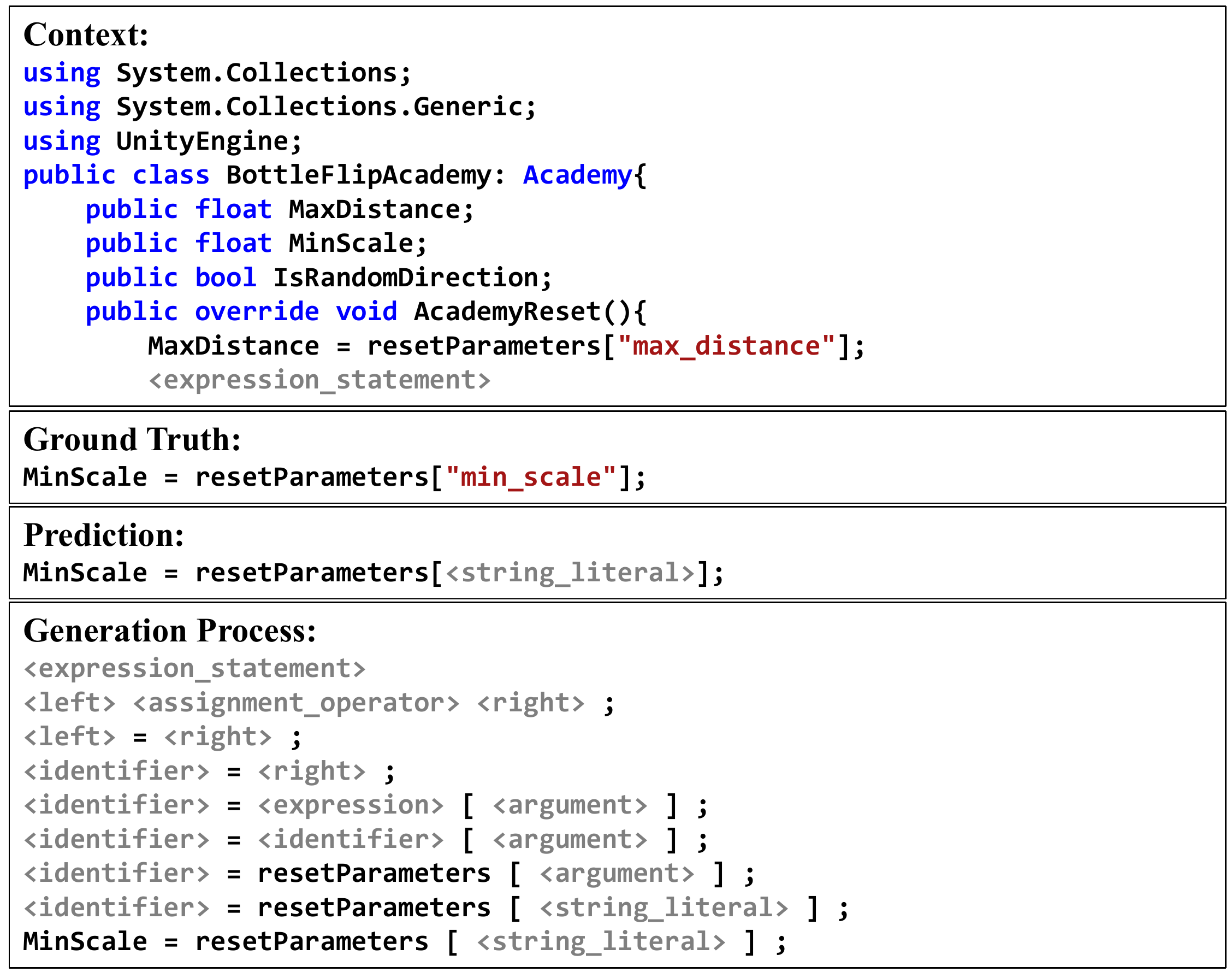}
		\caption{An example \projName generation for C\#. Each line in the generation process shows subsequent states of $\mathbf{x}_t$ in
		\autoref{alg:generation}. Here, \projName predicts a sketch that
		matches the ground-truth expansion, but places a hole at the key of the dictionary lookup,
		instead of predicting a low-likelihood string literal.}
		\label{fig:charp-generation-2}
	\end{center}
\end{figure}

\begin{figure}[h]
	\begin{center}
		\includegraphics[width=1\columnwidth]{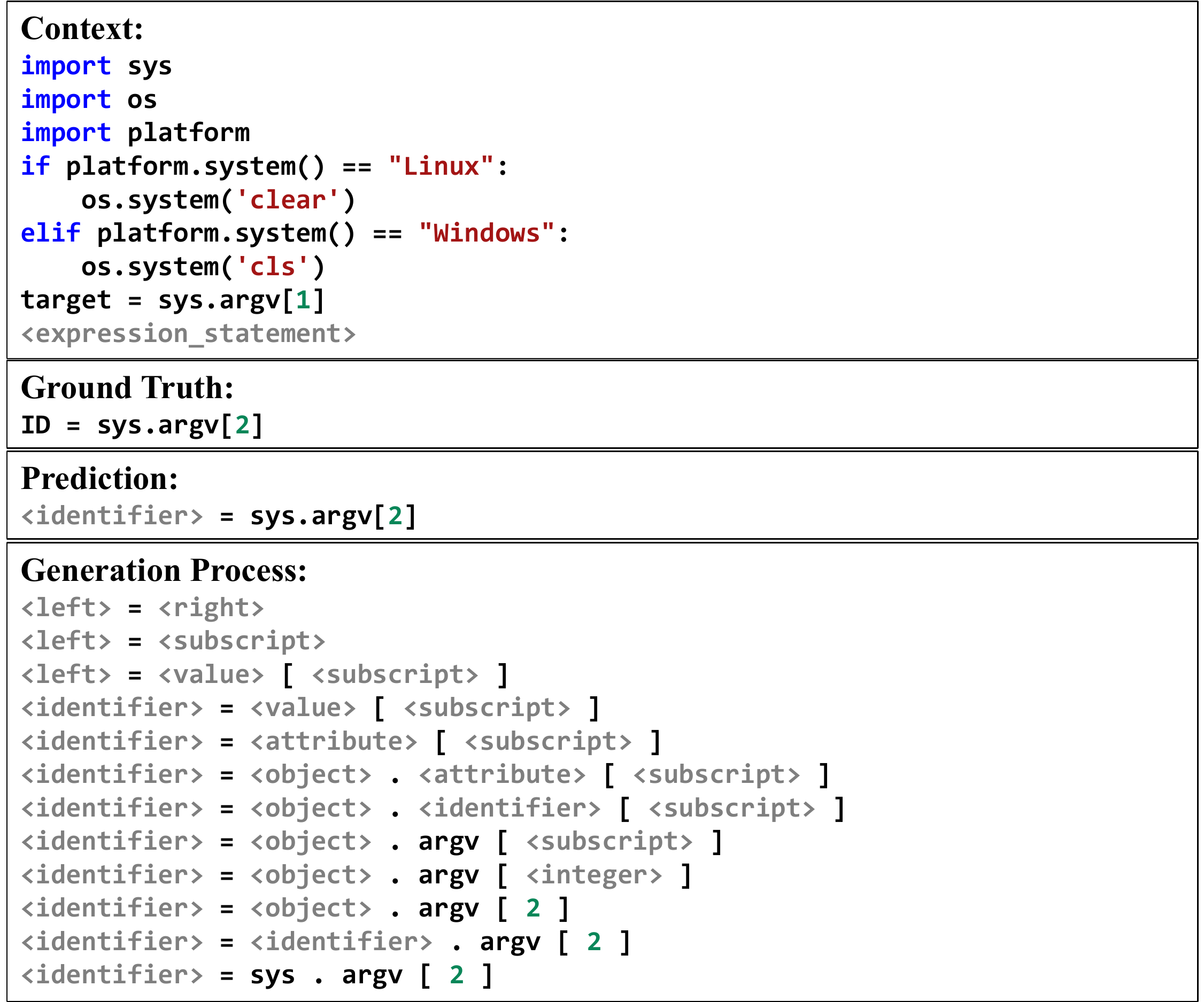}
		\caption{An example \projName generation for Python. Each line in the generation process shows subsequent states of $\mathbf{x}_t$ in
		\autoref{alg:generation}.
	    \projName here predicts that the user's intent is to read-in a second
	    argument and store it in a variable. However, within the current context,
	    the name of the variable storing the second argument would be impossible
	    to predict. \projName --- reasonably --- places a hole at the given location and generates a matching sketch. In this example, any traditional left-to-right model would need to first predict an accurate target variable name (which seems unlikely in the given context) before predicting the right-hand side of the assignment.
		}
		\label{fig:python-generation-1}
	\end{center}
\end{figure}

\begin{figure}[h]
	\begin{center}
		\includegraphics[width=1\columnwidth]{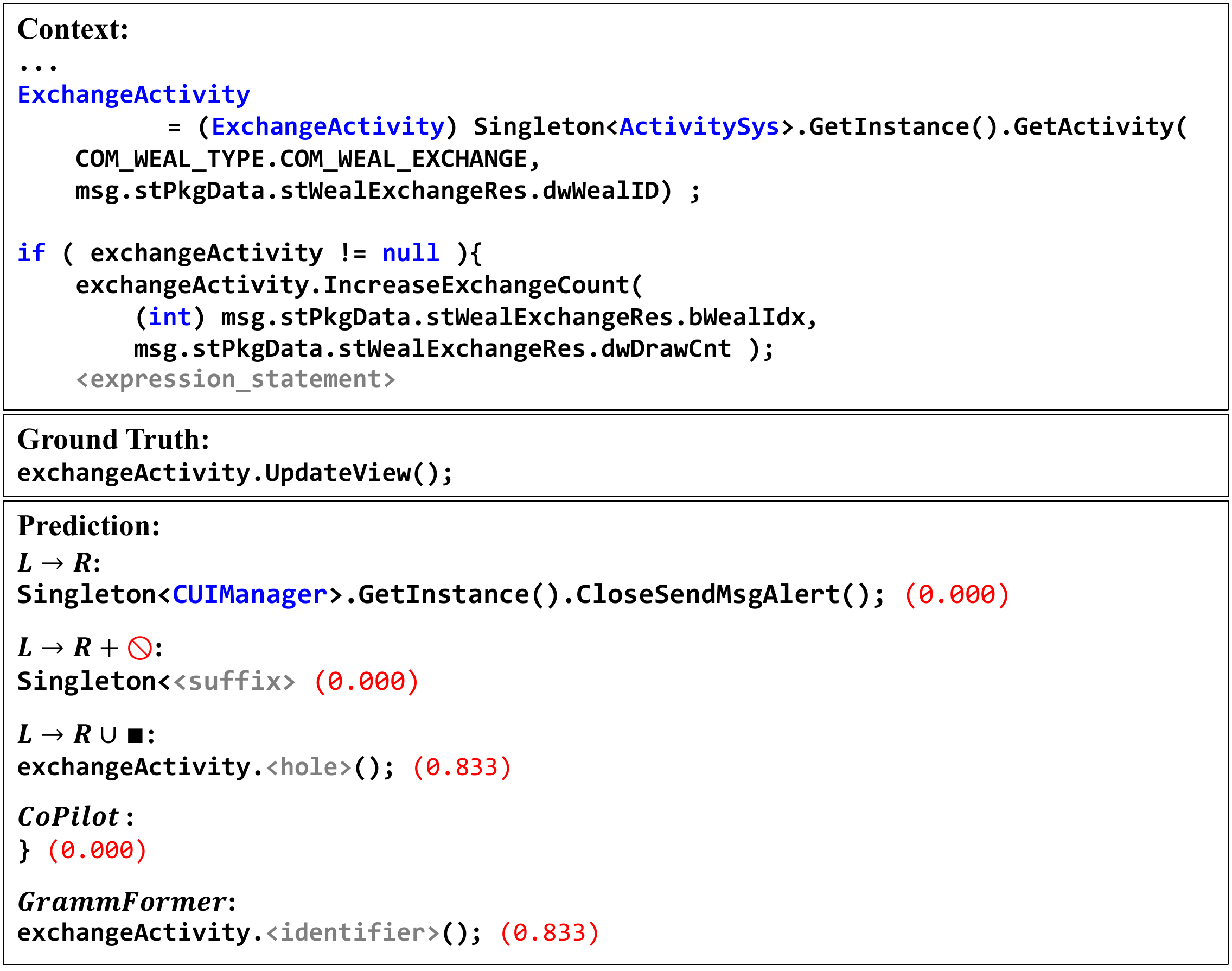}
		\caption{A C\# example and completion outputs from different models. \regexAcc score reported in {\color{red}red}. Here, \projName correctly identifies that 
		a method should be invoked on \texttt{exchangeActivity}, but does not predict the concrete method. If \projName was extended with information from a static analysis about the \texttt{ExchangeActivity} (potentially a user-defined type) then an accurate suggestion could have potential been made.}
		\label{fig:csharp-outputs-1}
	\end{center}
\end{figure}

\begin{figure}[h]
	\begin{center}
		\includegraphics[width=1\columnwidth]{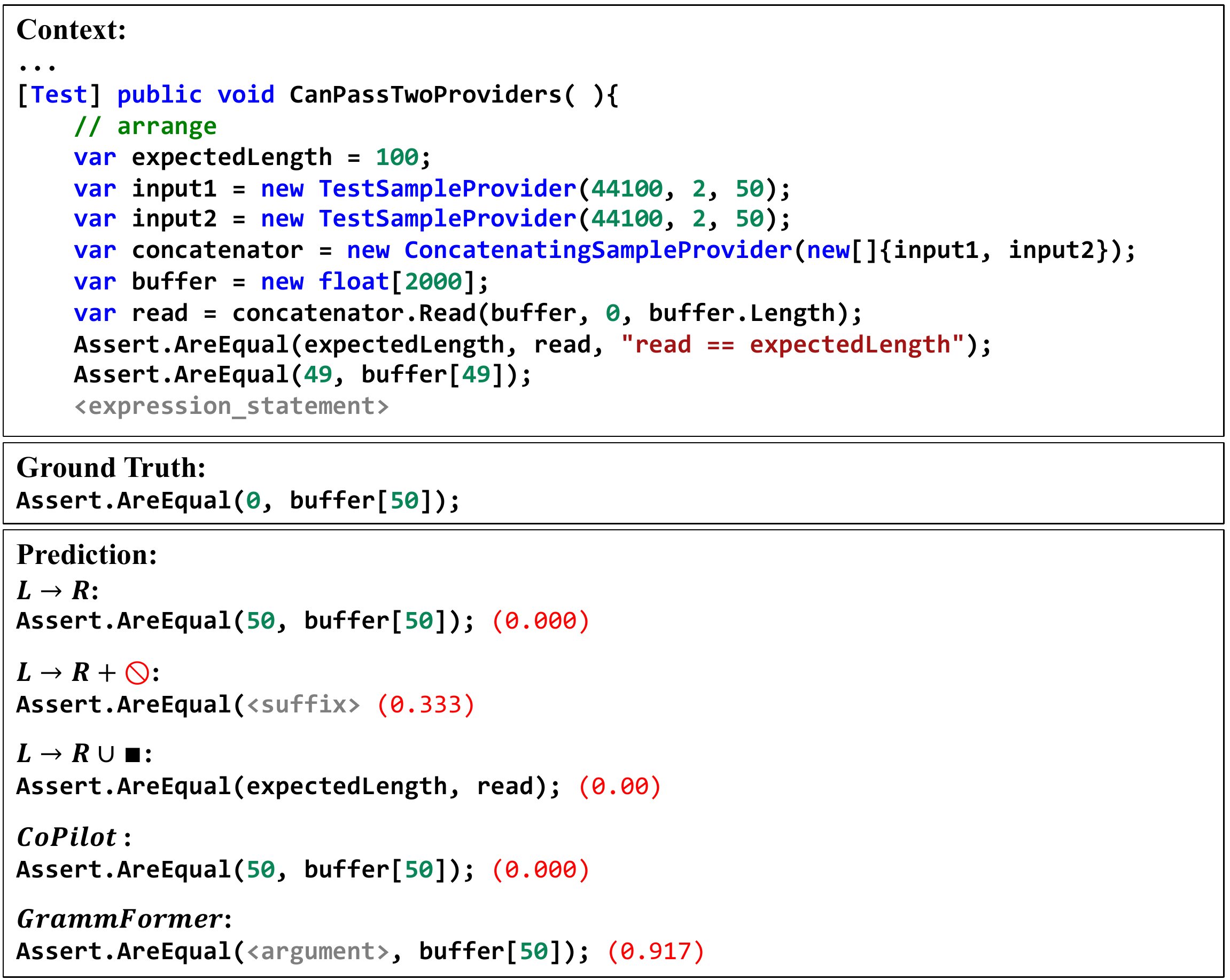}
		\caption{A C\# example and completion outputs from different models. \regexAcc score reported in {\color{red}red}. Here, \projName correctly predicts that an \texttt{AreEqual} assert statement should be made, checking the value of \texttt{buffer[50]}. However, within this context, the correct concrete expected value (\texttt{0}) would be hard to predict, even for a human. \projName places a hole there and generates a correct line-level sketch. In contrast, \lrbaseline introduces a wrong completion and \lrstopbaseline creates a correct, but much shorter sketch.
		}
		\label{fig:csharp-outputs-2}
	\end{center}
\end{figure}

\begin{figure}[h]
	\begin{center}
		\includegraphics[width=1\columnwidth]{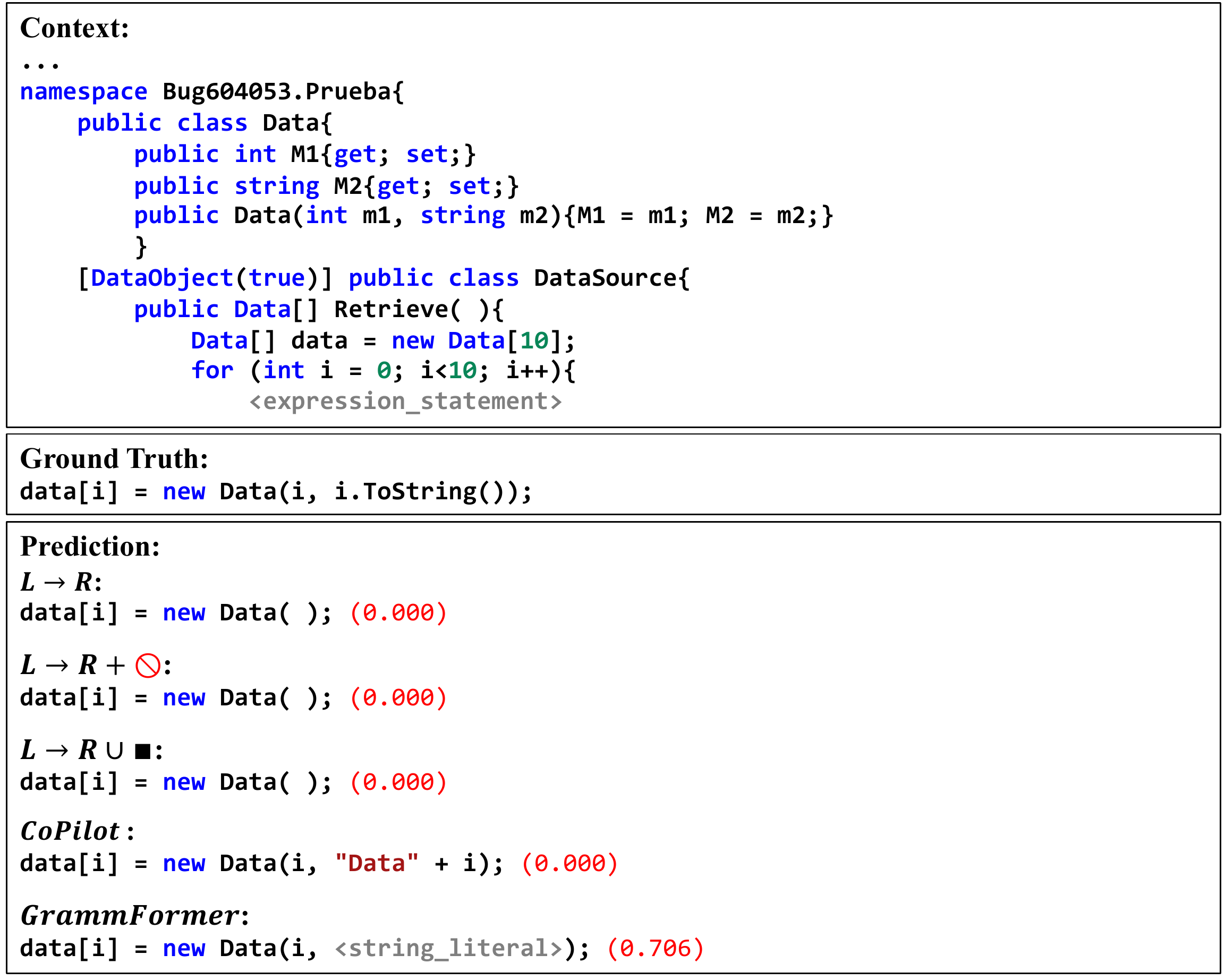}
		\caption{A C\# example and completion outputs from different models. \regexAcc score reported in {\color{red}red}. While all models predict that an assignment needs to be made to each \texttt{data[i]}, the exact form of the constructor is hard to predict. \projName seems to be looking at the constructor definition and predicts that some \NT{StringLiteral} needs to be used as the second argument, although it is uncertain about its concrete form, hence introducing a hole.}
		\label{fig:csharp-outputs-3}
	\end{center}
\end{figure}

\begin{figure}[h]
	\begin{center}
		\includegraphics[width=1\columnwidth]{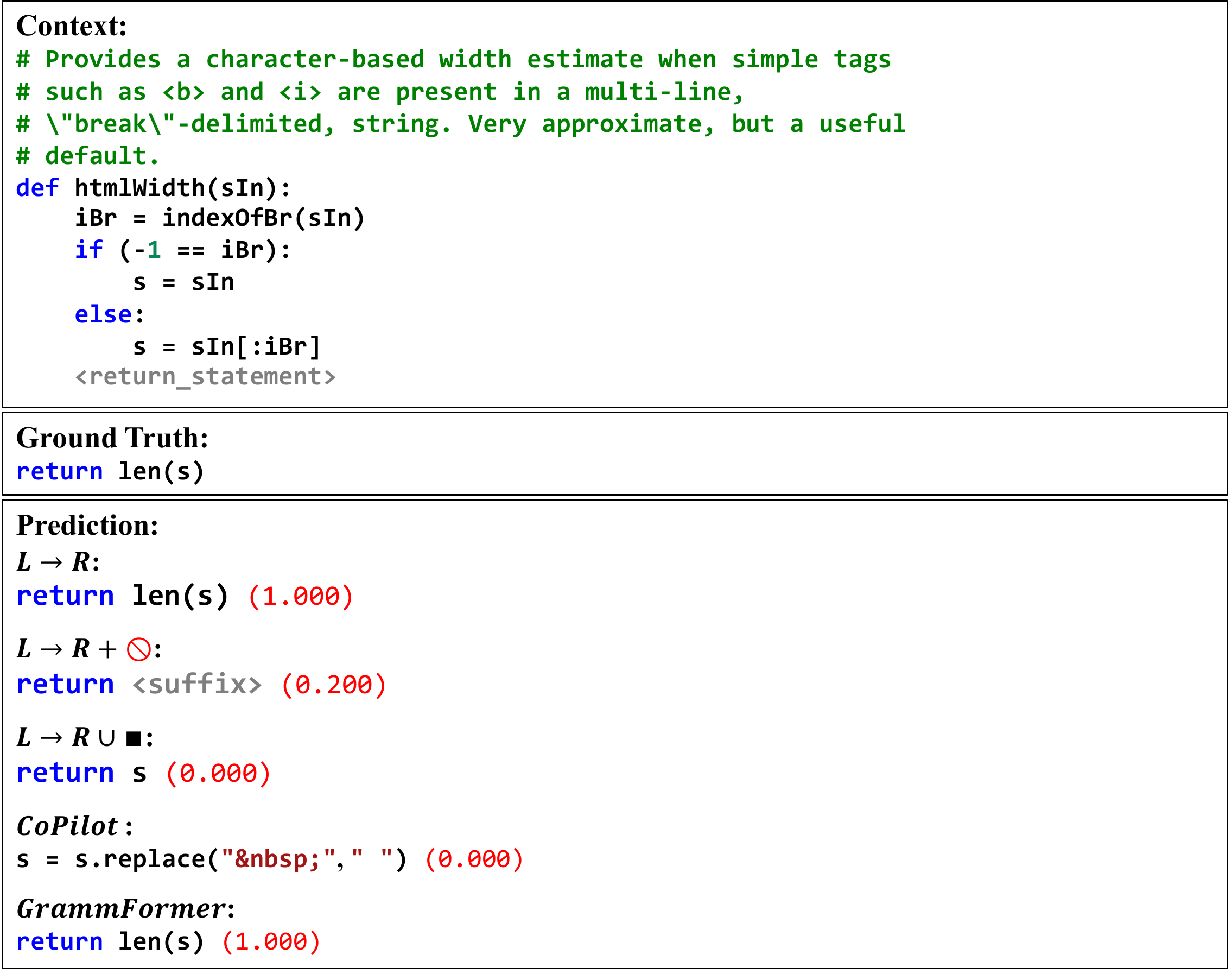}
		\caption{A Python example and completion outputs from different models. \regexAcc score reported in {\color{red}red}. Here both \lrbaseline and \projName predict the full line correctly, but \lrstopbaseline seems to return a more conservative (but correct) sketch.}
		\label{fig:python-outputs-1}
	\end{center}
\end{figure}

\begin{figure}[h]
	\begin{center}
		\includegraphics[width=1\columnwidth]{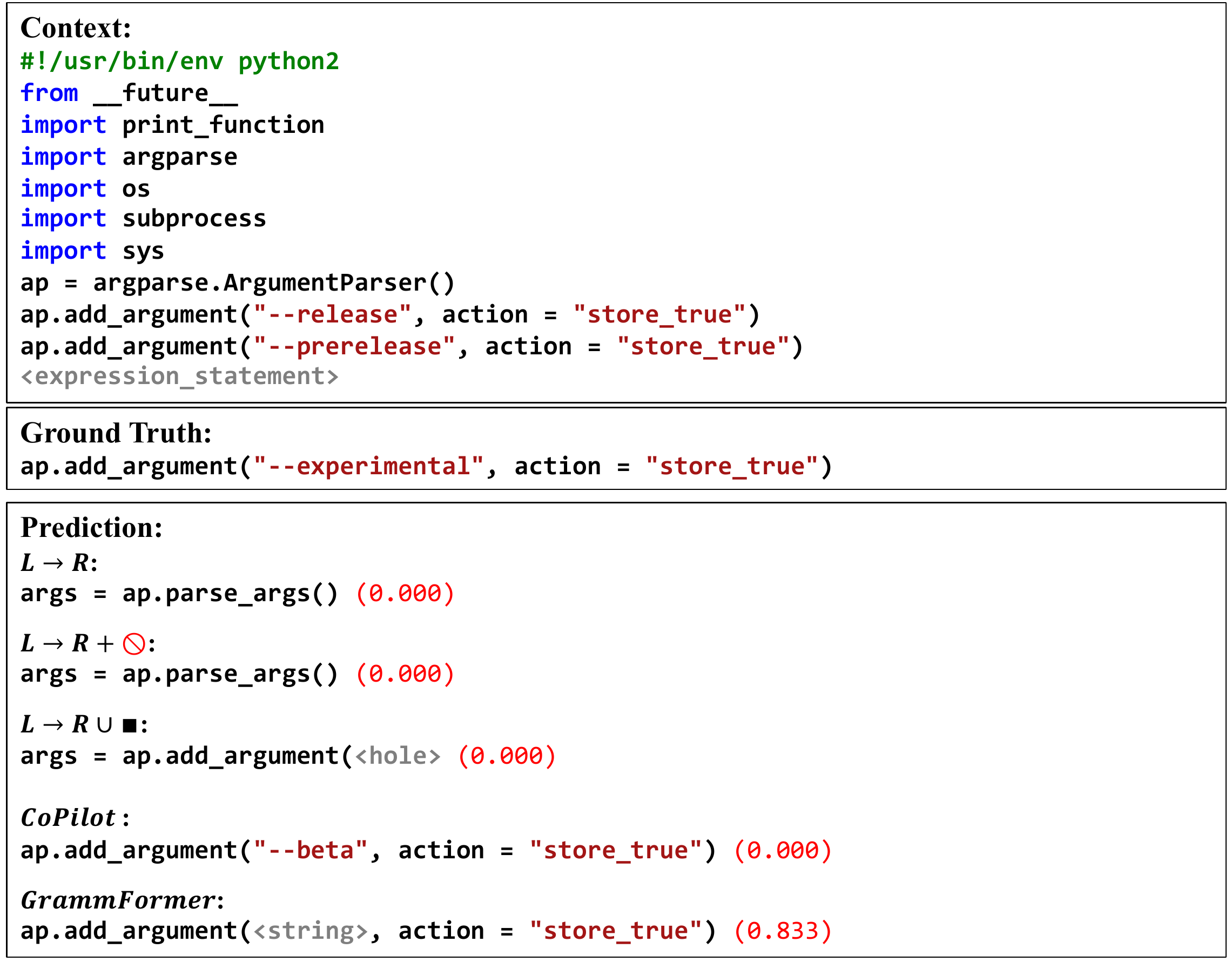}
		\caption{A Python example and completion outputs from different models. \regexAcc score reported in {\color{red}red}. See main text in the introduction for a description.}
		\label{fig:python-outputs-2}
	\end{center}
\end{figure}

\begin{figure}[h]
	\begin{center}
		\includegraphics[width=1\columnwidth]{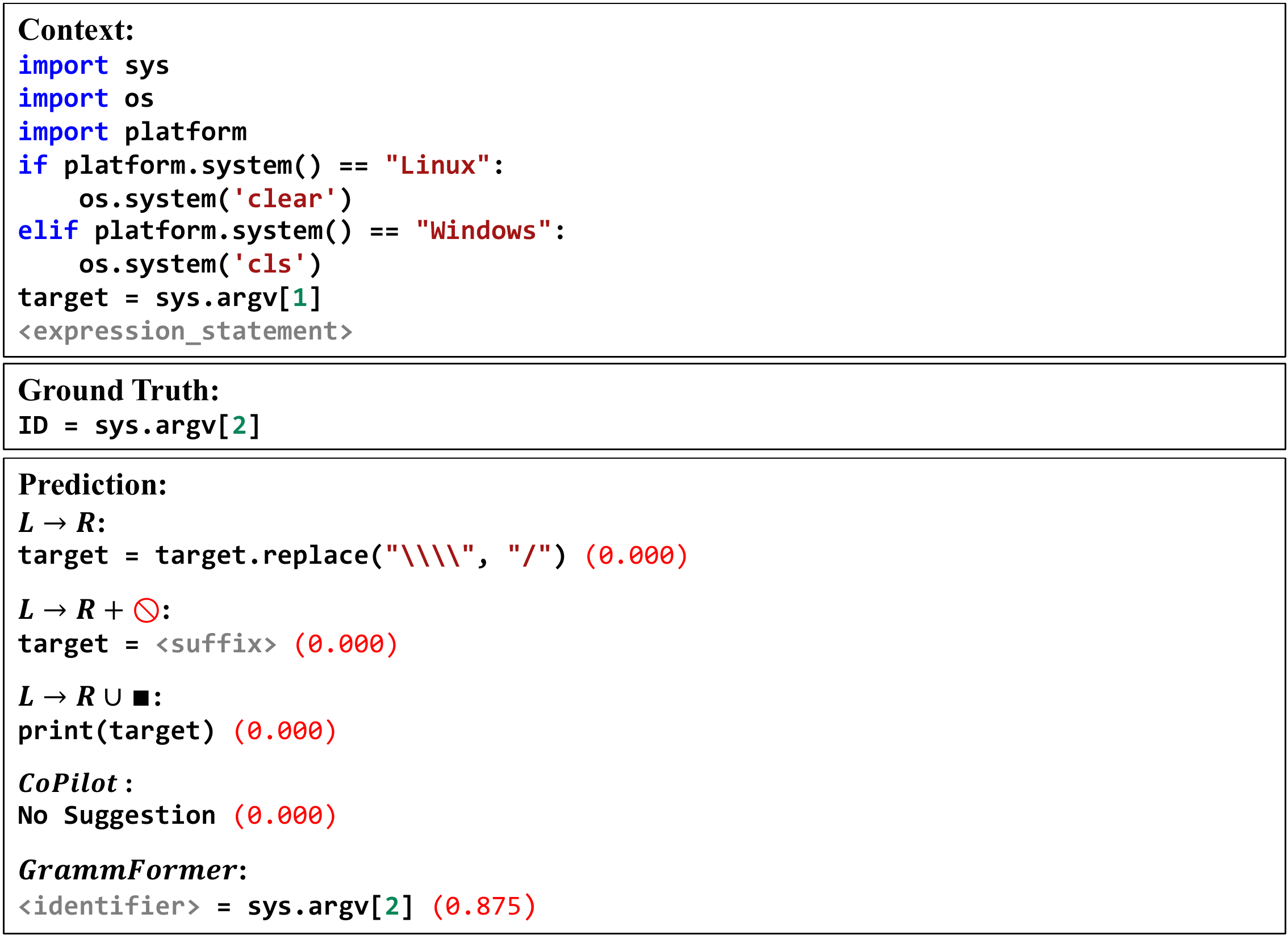}
		\caption{A Python example and completion outputs from different models. \regexAcc score reported in {\color{red}red}. Generation steps of \projName shown in \autoref{fig:python-generation-1}. \lrbaseline and \lrstopbaseline cannot generate correct sketches since the first token would be impossible to guess within this code context.}
		\label{fig:python-outputs-3}
	\end{center}
\end{figure}

\begin{figure}[h]
	\begin{center}
		\includegraphics[width=1\columnwidth]{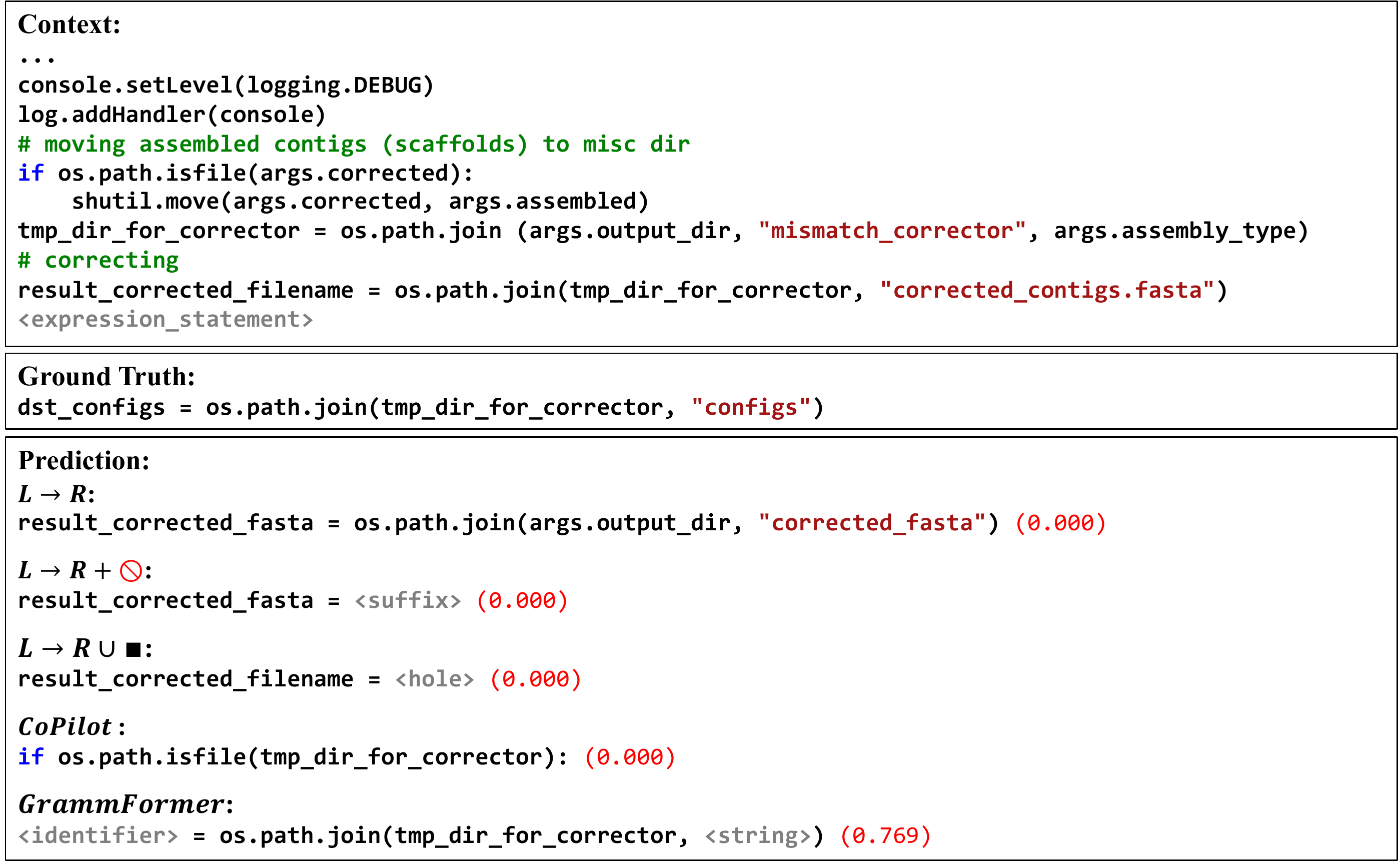}
		\caption{A Python example and completion outputs from different models. \regexAcc score  reported in {\color{red}red}. \projName completes the line creating a correct sketch with two holes at locations avoiding to make the mistakes that \lrbaseline and \lrstopbaseline makes.}
		\label{fig:python-outputs-4}
	\end{center}
\end{figure}

\begin{figure}[h]
	\begin{center}
		\includegraphics[width=1\columnwidth]{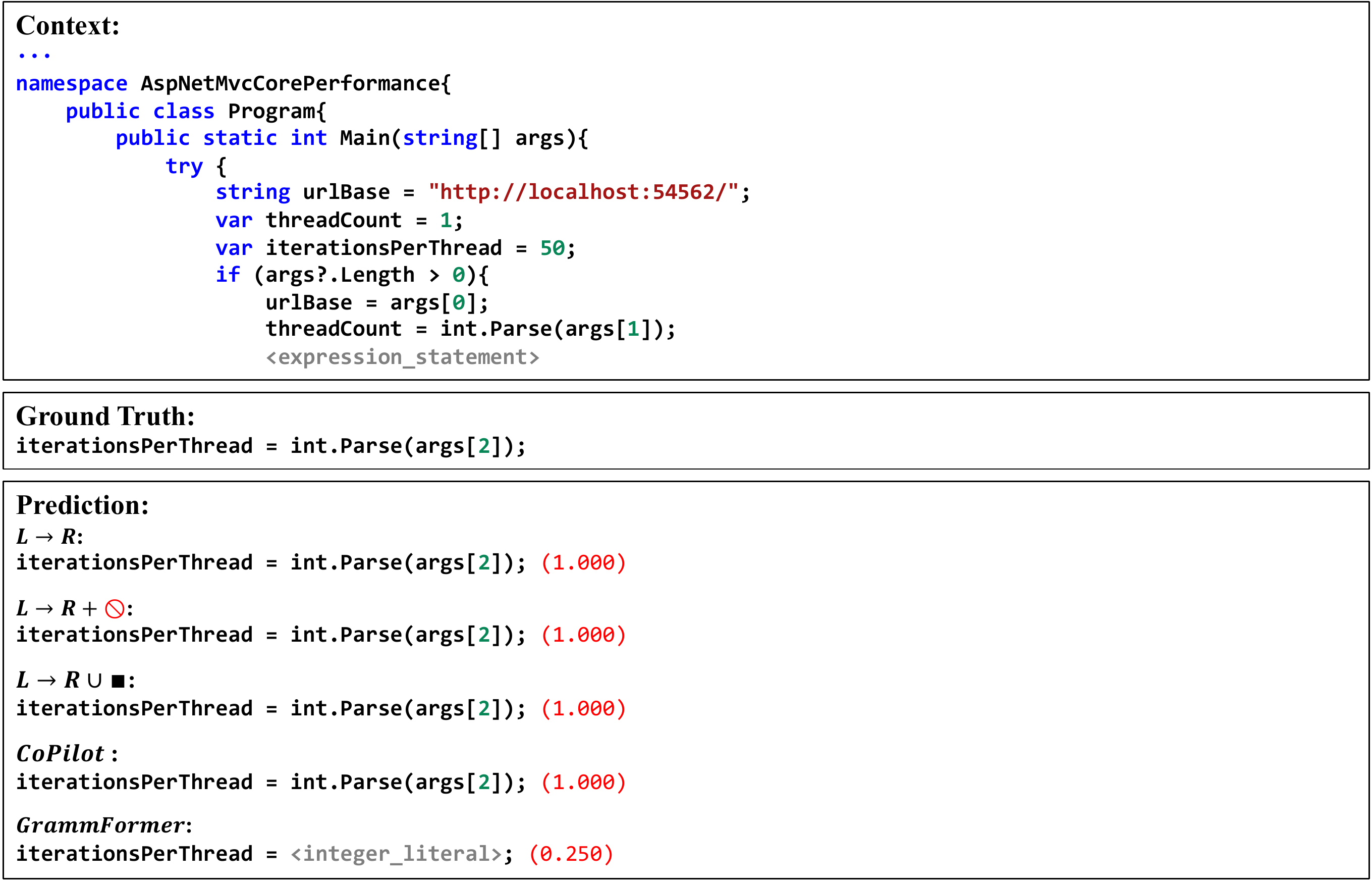}
		\caption{A C\# example and completion outputs from different models. \regexAcc score  reported in {\color{red}red}. \projName suggests a correct sketch but the right-hand side of the assignment has to stop expansion since \NT{IntegerLiteral} cannot generate \texttt{int.Parse(args[2])}. This suggests some of the limitations that the grammar-based generation of \projName may have, especially for shorter sequences.} 
		\label{fig:csharp-bad-case-2}
	\end{center}
\end{figure}

\begin{figure}[h]
	\begin{center}
		\includegraphics[width=1\columnwidth]{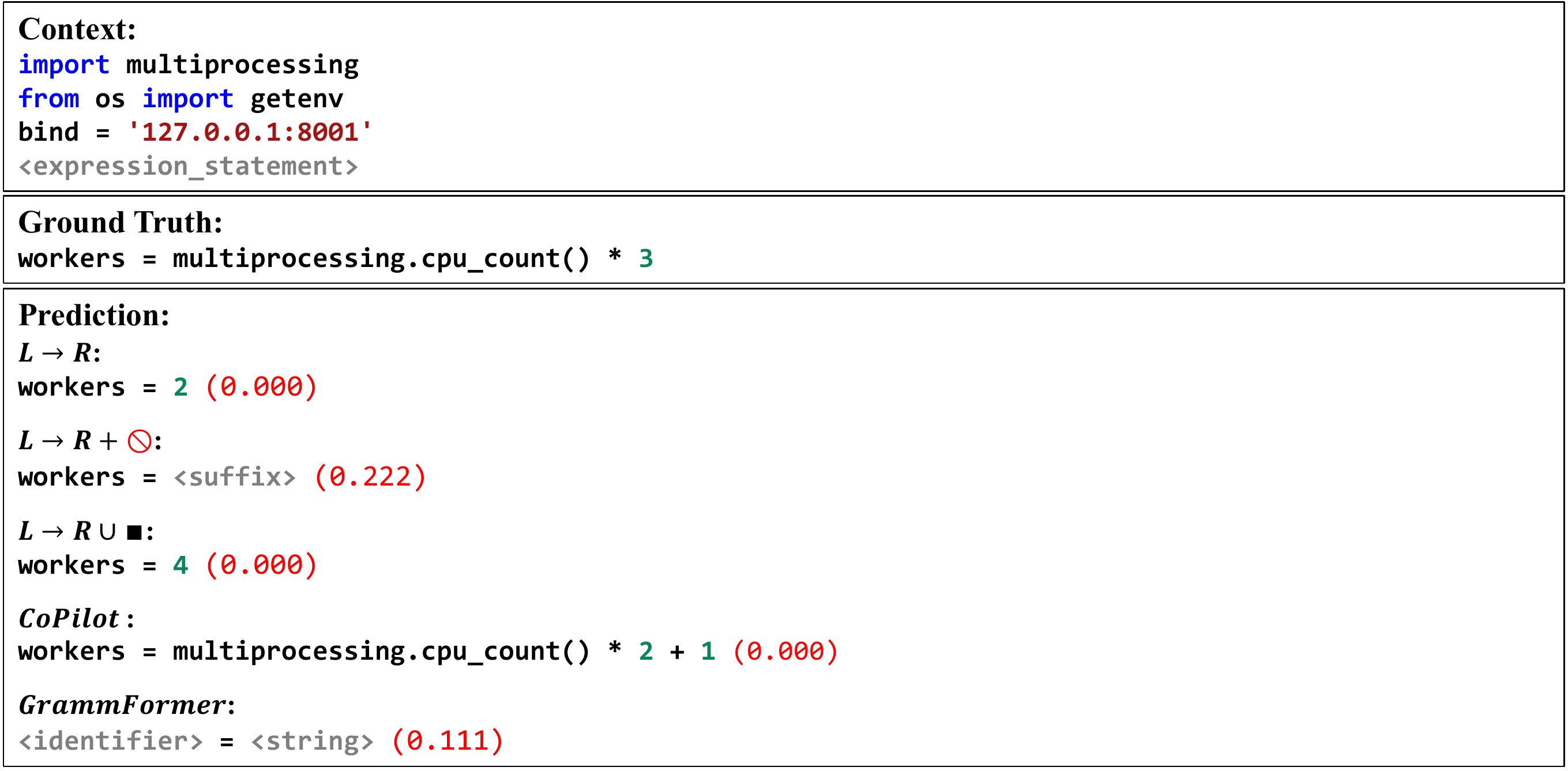}
		\caption{A Python example and completion outputs from different models. \regexAcc score  reported in {\color{red}red}. Although the sketch of the prediction from \projName is typically correct, it is not useful. Researching better evaluation metrics may improve \projName.} 
		\label{fig:python-bad-case-1}
	\end{center}
\end{figure}

\begin{figure}[h]
	\begin{center}
		\includegraphics[width=1\columnwidth]{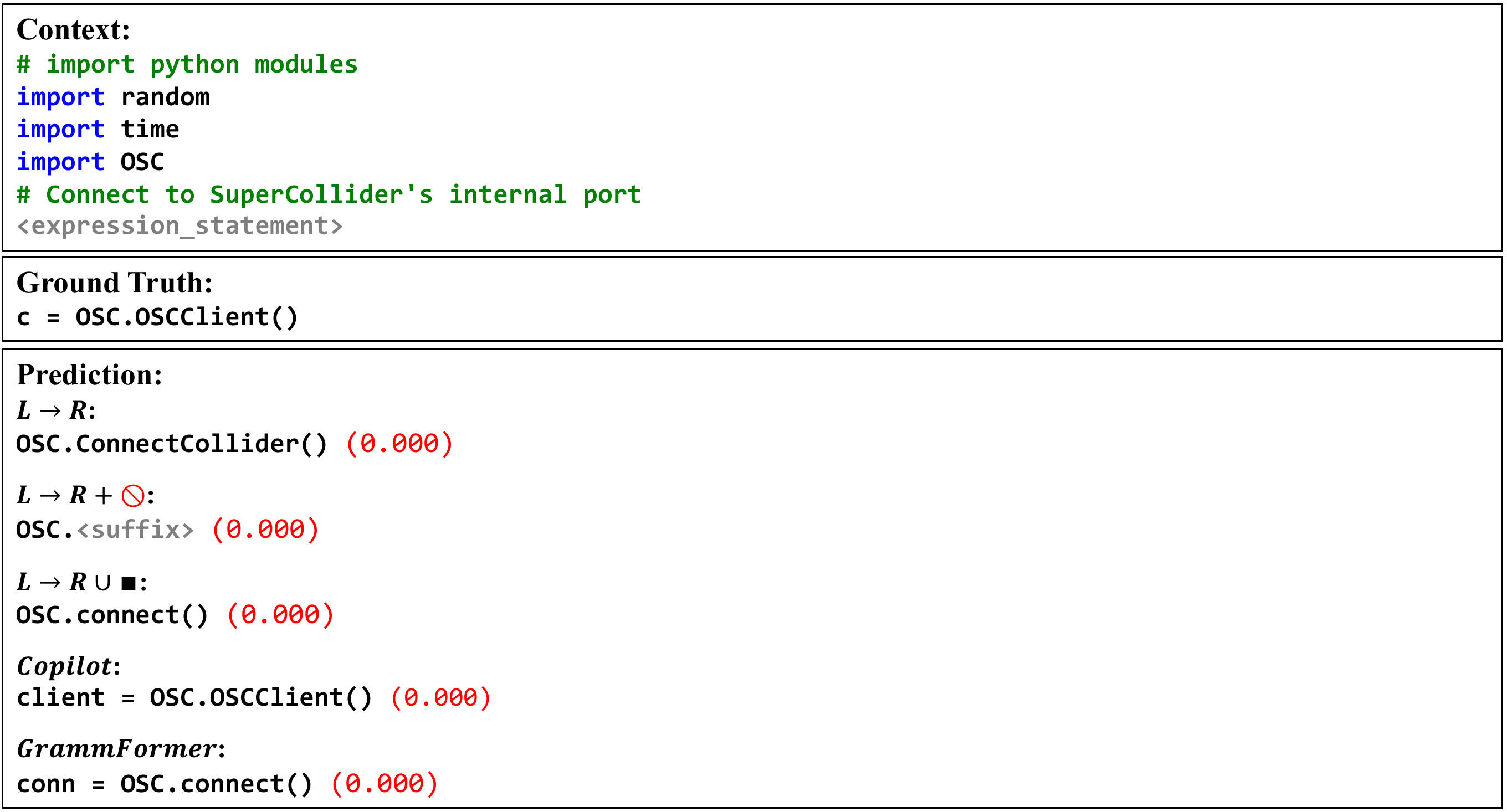}
		\caption{A Python example and completion outputs from different models. \regexAcc score  reported in {\color{red}red}. All model fail to invoke the correct API of the library. A potential future direction to mitigate the problem is to incorporate definitions of the external or system classes.} 
		\label{fig:python-bad-case-2}
	\end{center}
\end{figure}

\autoref{fig:charp-generation-2} and \autoref{fig:python-generation-1}
show two examples from our dataset along with the ground-truth and the sequence of expansions performed by \projName. \autoref{fig:csharp-outputs-1}-\ref{fig:python-outputs-4} show example generations by \projName and the baseline models \lrbaseline and \lrstopbaseline.
The parentheses in red indicate the \regexAcc score for each suggestion. For
the \lrstopbaseline baseline the special non-terminal \texttt{<suffix>} is added to indicate that a hole is introduced at the end of the left-to-right generation.
Finally \autoref{fig:csharp-bad-case-2}-\ref{fig:python-bad-case-2} show example generations where \projName make mistakes. A discussion for each of those sample is found at the caption of each figure.

\begin{figure}[h]
	\begin{center}
		\includegraphics[width=1\columnwidth]{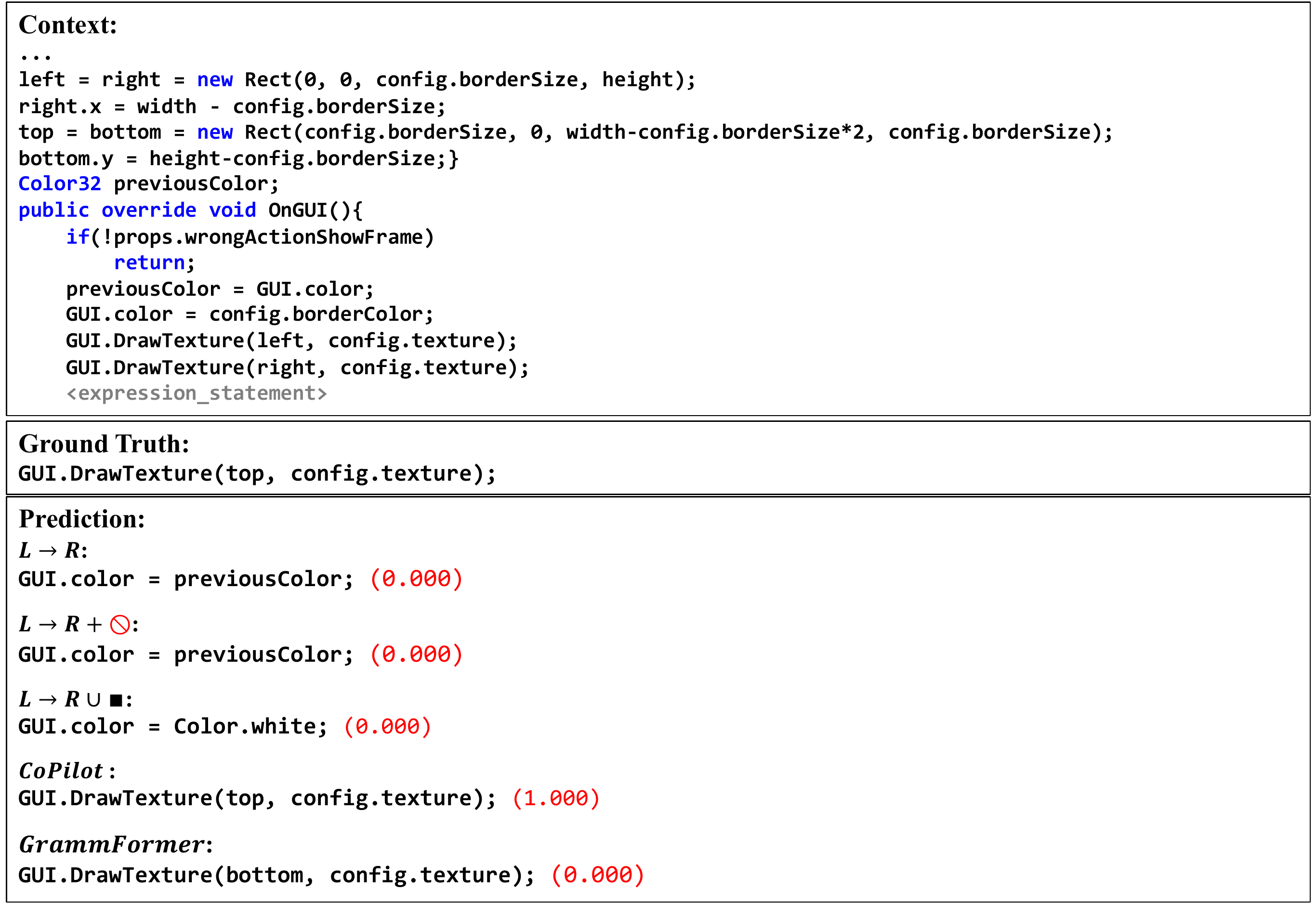}
		\caption{A C\# example and incorrect completion outputs from different models. \regexAcc score  reported in {\color{red}red}. The prediction from \projName is almost right but should have created a hole at the first argument for the user to fill-in. This shows that improved methods for training the policy network may improve results in the future.}
		\label{fig:csharp-bad-case-1}
	\end{center}
\end{figure}

\section{Dataset Statistics}
\label{appx:dataset stats}




Some statistics about the datasets used throughout this work
are shown in \autoref{tbl:dataset stats}
\begin{table}[th]
    \centering
    \begin{tabular}{lrr}\toprule
                                      & Python & C\# \\ \midrule
Num Training Files/Trees                  & 1973400  & 1948516 \\
Num Validation Files/Trees              & 218398   & 216299\\
Num Test Files/Trees                      & 460874   & 480166\\
Avg num tokens of $\mathbf{x}_t$          & 194.5    & 201.4\\
Median num tokens of $\mathbf{x}_t$       & 205      & 206 \\
99 percentile num tokens of $\mathbf{x}_t$& 250      & 260 \\
Avg num tokens of $\mathbf{y}$            & 1.9      & 1.9 \\
Median num tokens of $\mathbf{y}$         & 1        & 1\\
99 percentile num tokens of $\mathbf{y}$  & 9      & 7 \\
\bottomrule
    \end{tabular}
    \caption{Statistics of the datasets used.}
    \label{tbl:dataset stats}
\end{table}

\section{Flattened Non-Terminals}
\label{appx:expansions}
The non-terminals in \autoref{tbl:expanded nonterminals} are always expanded and are not considered as non-terminals. Most of these non-terminals have always the same children (terminals or non-terminals), representing a single CFG rule. By flattening those non-terminals the depth of tree is reduced (and hence the number of loops needed in \autoref{alg:generation}).
\begin{table}
\begin{tabular}{lp{10cm}}
Python & 
\lstinline|block|, \lstinline|tuple|, \lstinline|and|, \lstinline|or|, \lstinline|+|, \lstinline|-|, \lstinline|*|, \lstinline|/|, \lstinline|&|,
\lstinline+||+, \lstinline|//|, \lstinline|%|, \lstinline|@|, \lstinline|+=|, 
\lstinline|-=|, \lstinline|*=|, \lstinline|/=|, \lstinline|//=|, \lstinline|@=|, 
\lstinline|&=|, \lstinline+|=+, \lstinline|call|, \lstinline|keyword_argument|, \lstinline|name|, 
\lstinline|binary_operator|, \lstinline|for_in_clause|, \lstinline|unary_operator|, \lstinline|**|, \lstinline|true|, 
\lstinline|not_operator|, \lstinline|none|, \lstinline|false|, \lstinline|boolean_operator|, \lstinline|augumented_assignment|, 
\lstinline|await|, \lstinline|>>|, \lstinline|pair|, \lstinline+|+, \lstinline|parameters|, \lstinline|<<|,  
\lstinline|dictionary_comprehension|, \lstinline|ellipsis|, \lstinline|arguments|, \lstinline|assignment|, \lstinline|^|, \lstinline|~|
\\
C\#    & 
\lstinline|block|, \lstinline|tuple|, \lstinline|and|, \lstinline|or|,
\lstinline|+|, \lstinline|-|, \lstinline|*|, \lstinline|/|,
\lstinline|&|, \lstinline+||+, \lstinline|//|, \lstinline|%|,
\lstinline|@|, \lstinline|+=|, \lstinline|-=|, \lstinline|*=|,
\lstinline|/=|, \lstinline|//=|, \lstinline|%=|, \lstinline|@=|,
\lstinline|&=|, \lstinline+|=+, \lstinline|**|, \lstinline|>>|,
\lstinline+|+, \lstinline|<<|, \lstinline|^|, \lstinline|~|,
\lstinline|assignment_expression|, \lstinline|invocation_expression|, \lstinline|arguments|, \lstinline|member_access_expression|,
\lstinline|try_statement|, \lstinline|catch_clause|, \lstinline|conditional_expression|, \lstinline|==|,
\lstinline|array_type|, \lstinline|rank|, \lstinline|base_expression|, \lstinline|conditional_access_expression|,
\lstinline|member_binding_expression|, \lstinline|initializer|, \lstinline|null_literal|, \lstinline|>|,
\lstinline|element_access_expression|, \lstinline|subscript|, \lstinline|??|, \lstinline|this_expression|,
\lstinline|implicit_array_creation_expression|, \lstinline|cast_expression|, \lstinline|!=|, \lstinline|variable_declaration|,
\lstinline|implicit_type|, \lstinline|&&|, \lstinline|as_expression|, \lstinline|as|,
\lstinline|<|, \lstinline|local_declaration_statement|, \lstinline|if_statement|, \lstinline|>=|,
\lstinline|<=|, \lstinline|throw_expression|, \lstinline|default_expression|, \lstinline|pattern|,
\lstinline|is_pattern_expression|, \lstinline|binary_expression|, \lstinline|bracketed_argument_list|, \lstinline|name|,
\lstinline|object_creation_expression|, \lstinline|await_expression|, \lstinline|,|
\end{tabular}
\caption{Non-terminals that are always expanded in the Tree-Sitter grammar for the two languages considered.}\label{tbl:expanded nonterminals}
\end{table}

\section{Understanding \regexAcc}
Since \regexAcc is a new metric, we
include two deterministic ways of introducing sketches in \autoref{tbl:regexacc baselines}. First, if all literals (strings, numeric) are replaced with a hole, we see that a high \regexAcc is achieved. In contrast, replacing both identifiers and literals (leaving ``just'' parentheses, brackets, dots, \etc) we get an easy ``lower-bound''. Note how C\# --- which is 
syntactically more verbose ---  achieves a better score, compared to Python.
\begin{table}[htb]
    \centering
    \begin{tabular}{lrr} \toprule
 & C\# & Python \\ \midrule
Replace all literals with holes & 0.865 & 0.608\\
Replace all literals and identifiers with holes& 0.126 & 0.060 \\ \bottomrule 
    \end{tabular}
    \caption{\regexAcc when deterministically introducing holes at specific location.}
    \label{tbl:regexacc baselines}
\end{table}
In \autoref{tbl:regex acc samples}, we show some example sketches and their associated \regexAcc score.

\begin{table}[h]\centering
\caption{Example \regexAcc scores for a variety of sketches.}\label{tbl:regex acc samples}
\begin{tabular}{lr} \toprule
Ground-truth \\
\lstinline|ap.add_argument("--experimental", action="store_true")| \\ \midrule
& \regexAcc \\ \cmidrule{2-2}
\lstinline|ap.add_argument(|\hole\lstinline|, action="store_true")| & 0.9 \\
\lstinline|ap.add_argument(|\hole\lstinline|, action=|\hole\lstinline|)|& 0.8\\
\lstinline|ap.add_argument(|\hole\lstinline|, |\hole\lstinline|)|& 0.6\\
\lstinline|ap.add_argument(|\hole\lstinline|, action=|{\color{red}\lstinline|"store_false"|}\lstinline|)| & 0.0 \\
\lstinline|ap.add_argument(|\hole\lstinline|, |{\color{red}\lstinline|required=|\hole}\lstinline|)| & 0.0 \\\bottomrule
\end{tabular}
\end{table}

\section{Beam search}\label{appx:beam search}
\autoref{alg:beam search} presents the beam search used in \projName. 
\begin{algorithm}[h]
    \caption{\projName beam search, given an input sequence $\mathbf{x}_0$.}\label{alg:beam search}
\begin{algorithmic}
\State{$b \leftarrow \{(\mathbf{x}_0, 0, \texttt{false})\}$} \Comment{Initialize Beam (state, logprob, isDone)}
\While{$\exists (\mathbf{x}, p, isDone) \in b \text{~with~} isDone = \texttt{false}$ } \Comment{While beam contains incomplete generations}
    \State{$b' \leftarrow \{\}$
    \For{$(\mathbf{x}, p, isDone) \in b$}\Comment{For each sample in beam}
        \If{isDone}  \Comment{If suggestion is complete}
            \State{$b' \leftarrow b' \cup \{(x, p, isDone)\}$} \Comment{No operation, beam is complete}
            \State{\textbf{continue}}
        \EndIf
        \For{$i \in \textsc{TopM}(P_s(i \vert \mathbf{x}, N(\mathbf{x})))$} \Comment{Get top-$m$ non-terminal positions}
            \State{$p_s \leftarrow \log P_s(i\vert \mathbf{x}, N(\mathbf{x}))$}
            \If{$i = \stopExpand$}
                \State{$b' \leftarrow b' \cup \{(x, p + p_s, \texttt{true})\}$} \Comment{Stop Expansion}
            \Else
                \For{$\mathbf{y} \in \textsc{TopN}(P_e(\mathbf{y}\vert\mathbf{x},i)$)}\Comment{Beam search on $\mathbf{y}$ yields $n$ candidates}
                    \State{$p_e \leftarrow P_e(\mathbf{y}\vert\mathbf{x},i)$}
                    \State{$b' \leftarrow b' \cup \{(\mathbf{x}_{<i}::\mathbf{y}::\mathbf{x}_{>i}), p + p_s + p_e, \texttt{false})\}$}\Comment{Expand $x_i$}
                \EndFor
            \EndIf
        \EndFor
    \EndFor
    \State{$b\leftarrow \textsc{TopK}(b')$} \Comment Prune Candidates and keep top $k$}
\EndWhile
\Return $b$
\end{algorithmic}
\end{algorithm}
\end{document}